\documentclass[runningheads]{llncs}

\usepackage{eccv}

\usepackage{eccvabbrv}

\usepackage{graphicx}
\usepackage{booktabs}
\usepackage{algorithm}
\usepackage{listings}
\usepackage{xcolor}
\usepackage{multirow}        
\usepackage[table]{xcolor}   
\usepackage{array}           

\usepackage[accsupp]{axessibility}  

\definecolor{commentcolor}{RGB}{110,154,155}
\newcolumntype{L}[1]{>{\raggedright\let\newline\\\arraybackslash\hspace{0pt}}m{#1}}
\newcolumntype{C}[1]{>{\centering\let\newline\\\arraybackslash\hspace{0pt}}m{#1}}
\newcolumntype{R}[1]{>{\raggedleft\let\newline\\\arraybackslash\hspace{0pt}}m{#1}}
\definecolor{LightGray}{RGB}{240,240,240}
\definecolor{codecolor}{RGB}{54,125,189}

\def\eg{\emph{e.g.}\xspace}

\usepackage{hyperref}

\usepackage{orcidlink}

\begin{document}

\title{Unbalanced Optimal Transport for Efficient Visual Document Retrieval} 



\author{
Hoyeon Shin\inst{1}$^{\star}$\orcidlink{0009-0007-8016-5022} 
\and
Jeongyeon Kim\inst{1,2}\thanks{Equal contribution.}\orcidlink{0009-0004-1520-6201} 
\and
Yeong Jun Koh\inst{3}\orcidlink{0000-0003-1805-2960}
\and 
Yeoneung Kim\inst{1}\orcidlink{0000-0001-6293-3939}
\and 
Hanul Kim\inst{1}\thanks{Corresponding author.}\orcidlink{0000-0001-7450-6600}}
\authorrunning{H.~Shin et al.}

\institute{Seoul National University of Science and Technology, Seoul, Korea
\email{\{hy\_shin,jy\_kim, yeoneung, hukim\}@seoultech.ac.kr}
\and
SIZL, Seongnam, Korea
\and
Chungnam National University, Daejeon, Korea
\email{\{yjkoh\}@cnu.ac.kr}
}

\maketitle

\begin{abstract}
  Vision-Language Models (VLMs) have advanced Visual Document Retrieval (VDR) by enabling OCR-free, multi-vector representations. However, these models generate excessive tokens to preserve dense visual details, which severely limits retrieval scalability. Existing compression methods mitigate this overhead but rely on isolated heuristics that fail to balance token importance, semantic redundancy, and coverage capacity. We propose a visual token compression framework formulated as an Unbalanced Optimal Transport (UOT) problem. Our approach unifies these three objectives into a single alternating optimization process. Furthermore, we theoretically show that minimizing the transport cost upper-bounds the MaxSim retrieval error, ensuring ranking stability. Extensive experiments on the ViDoRe V1 and V3 benchmarks demonstrate that our method preserves fine-grained document semantics and improves the efficiency-effectiveness trade-off over state-of-the-art baselines. Code is available at~\url{https://github.com/shhhhhyy/Unbalanced-Optimal-Transport-for-EVDR}.
\end{abstract}

\section{Introduction}\label{sec:intro}
Visual Document Retrieval (VDR) finds relevant document images for a given text query. Early VDR systems~\cite{doermann1998survey,mitra2000survey} relied on optical character recognition (OCR)~\cite{mori1992ocr, smith2007tesseract} pipelines to feed parsed page information to standard text retrievers. However, these pipelines incur high indexing latency and struggle on visually complex pages featuring dense tables, figures, and irregular layouts~\cite{faysse2025colpali}. Therefore, recent studies~\cite{ma2024unifying, faysse2025colpali} have shifted to OCR-free paradigms driven by Vision-Language Models (VLMs)~\cite{radford2021learning, liu2023visual, wang2024qwen2, bai2025qwen3}. By mapping visual documents into a unified embedding space, VLM-based retrievers bypass OCR bottlenecks and inherently preserve structural semantics.

Within VLM-based retrievers, encoding visual documents requires balancing search speed and fine-grained accuracy. Compressing a page into a single global vector enables fast retrieval but often sacrifices localized visual details critical for complex queries. To address this limitation~\cite{ma2024unifying}, multi-vector models like ColPali~\cite{faysse2025colpali} represent each page as $N$ patch-level visual tokens. They evaluate relevance through late interaction~\cite{khattab2020colbert}, using a MaxSim operation to match each query token with the most similar page token. However, retaining all visual tokens creates a severe scalability bottleneck. Because scoring $M$ query tokens against $N$ document tokens requires $\mathcal{O}(NM)$ operations, index size and query latency become too costly for large-scale corpora.

To mitigate this overhead, recent approaches~\cite{clavie2024reducing, ma2025towards, yan2025docpruner} compress the $N$ visual tokens into a compact set of $K$ representatives. Standard techniques include pruning, which discards tokens based on attention scores, and merging, which aggregates similar patches through clustering. Despite such efficiency, these methods typically treat token importance and semantic redundancy as separate objectives. They also overlook the coverage capacity of the merged tokens. Unconstrained merging often collapses too many distinct patches into a single vector, degrading the fine-grained diversity needed for late interaction. Therefore, effective compression must balance three objectives: minimizing semantic redundancy, retaining informative tokens, and regulating coverage capacity.

\begin{figure*}[!t]
    \centering
    \begin{subfigure}[!t]{0.495\linewidth}
        \centering
        \includegraphics[width=\linewidth]{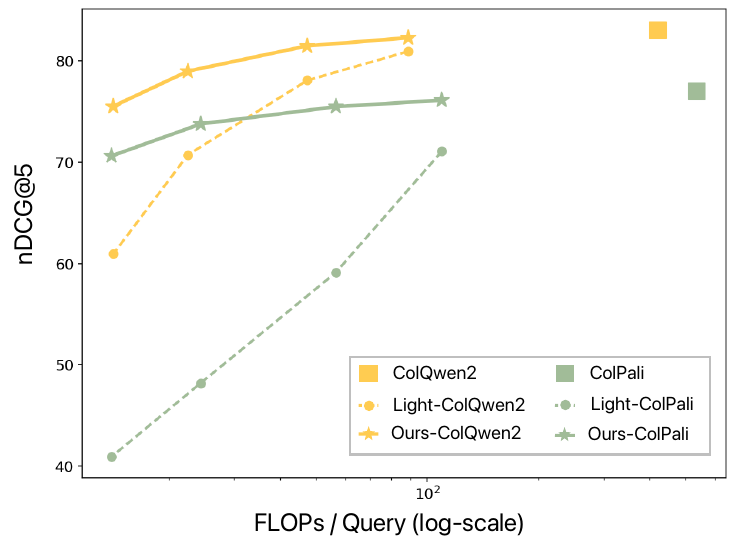}
        \caption{ViDoRe V1}
        \label{fig:flops_vidorev1}
    \end{subfigure}
    \hfill
    \begin{subfigure}[!t]{0.495\linewidth}
        \centering
        \includegraphics[width=\linewidth]{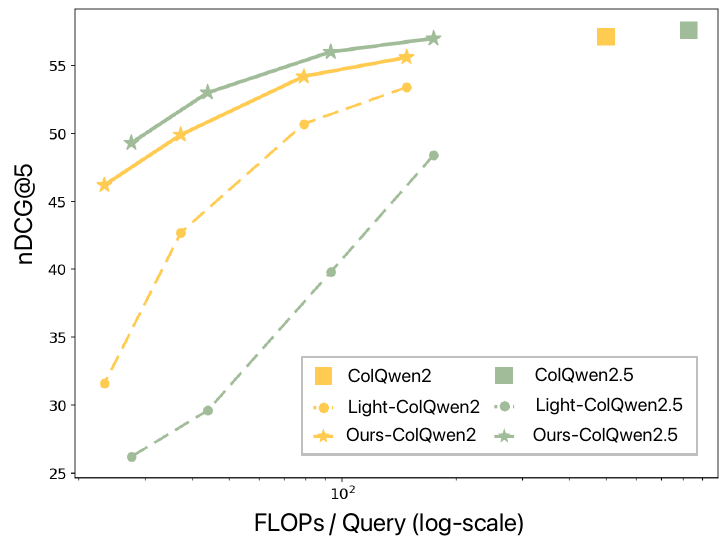}
        \caption{ViDoRe V3 English Tasks}
        \label{fig:flops_v3}
    \end{subfigure}
    \caption{\textbf{Efficiency--effectiveness trade-off.} nDCG@5 vs. FLOPs/query (log-scale) on ViDoRe benchmarks~\cite{faysse2025colpali, loison2026vidorev3}. FLOPs are estimated via a late-interaction proxy proportional to corpus size, expected query-document token interactions, and embedding dimension. Our method establishes a superior Pareto frontier compared to recent state-of-the-art baselines~\cite{faysse2025colpali, ma2025towards}.}
    \label{fig:flops}
\end{figure*}

In this paper, we propose a visual token compression framework for efficient VDR formulated as an Unbalanced Optimal Transport (UOT)~\cite{frogner2015learning, liero2018optimal, chizat2018scaling} problem. Our formulation naturally addresses three main objectives: the transport cost minimizes semantic redundancy by clustering similar patches, the source marginals prioritize informative tokens, and the target marginals regulate coverage capacity to prevent representation collapse. By solving this UOT problem through an efficient alternating optimization process, our method achieves a highly compact yet diverse set of representative tokens. In addition, we analyze the theoretical properties of the UOT objective and prove that its minimization effectively upper-bounds the MaxSim retrieval error. Extensive experiments on VDR benchmarks~\cite{faysse2025colpali, loison2026vidorev3} validate the effectiveness of our approach, delivering higher retrieval performance at the same compute and comparable accuracy, and demonstrating a superior Pareto frontier over recent state-of-the-art methods~\cite{faysse2025colpali, ma2025towards}, as depicted in~\cref{fig:flops}.

Our main contributions are summarized as follows:
\begin{itemize}
    \vspace{-0.2cm}
    \item We present a VDR-specific compression principle that balances token importance, semantic redundancy, and coverage capacity.
    \item We propose a visual token compression framework formulated as a UOT problem, unifying these three objectives into an efficient alternating optimization process.
    \item We provide rigorous theoretical guarantees proving that minimizing the unbalanced transport cost upper-bounds the MaxSim retrieval error degradation, alongside asymptotic and complexity analyses.
    \item Extensive experiments on standard VDR benchmarks demonstrate that our method consistently improves the efficiency--effectiveness trade-off compared to state-of-the-art baselines.
\end{itemize}

\section{Related Work}\label{sec:related}

\subsection{Visual Document Retrieval}
Visual Document Retrieval (VDR) aims to retrieve relevant document images from large-scale corpora. Early VDR systems~\cite{doermann1998survey, mitra2000survey} relied on OCR to convert images into text for standard retrieval. Although later layout-aware methods~\cite{katti2018chargrid, denk2019bertgrid, xu2020layoutlm} integrated spatial cues, their continued reliance on the OCR pipeline caused high indexing latency and difficulties parsing complex structures like tables and diagrams. These constraints led to OCR-free frameworks that leverage Vision-Language Models (VLMs) to encode pages directly. For instance, DSE~\cite{ma2024unifying} proposed representing document images as single vectors to prioritize efficiency, though this approach often lacks the localized detail needed for precise matching. In contrast, ColPali~\cite{faysse2025colpali} introduced a multi-vector late interaction~\cite{khattab2020colbert} to preserve these details, which yields higher accuracy. However, the resulting visual token explosion presents a significant bottleneck for storage and retrieval speed.

\subsection{Visual Token Compression}
To mitigate visual token overhead, various token reduction techniques have been proposed. General-purpose methods exploit semantic redundancy in vision transformers via pruning and merging~\cite{rao2021dynamicvit, yang2025visionzip}, or saliency-driven selection~\cite{chen2024image, zhang2025beyond}. Beyond simple pruning heuristics, FitPrune~\cite{ye2025fit} uses attention-distribution divergence for training-free pruning, while TopV~\cite{yang2025topv} estimates token contribution via inference-time transport optimization. However, document images are notably dense, where task-critical information—such as table cells and captions—is often concentrated in small regions, making aggressive reduction risky. In the visual document domain, DocKylin ~\cite{zhang2025dockylin} proposed efficient visual slimming at both the pixel and token levels to handle high-resolution, text-dense pages. To address this, VDR-specific efforts have explored clustering-based pooling~\cite{clavie2024reducing}, semantic clustering~\cite{ma2025towards}, and attention-based pruning~\cite{yan2025docpruner}. Despite these advancements, existing approaches typically rely on a single heuristic—similarity or importance—and often fail to control merge imbalance. Such imbalance can lead to token over-collapse, ultimately degrading the coverage required for effective late interaction. Orthogonal to these methods, two-stage pipelines~\cite{kim2025hybrid} balance efficiency by combining filtering with multi-vector reranking.

\subsection{Optimal Transport in Computer Vision}
Optimal Transport (OT)~\cite{villani2009optimal, peyre2019computational, montesuma2024recent} provides a principled framework for aligning distributions by minimizing the cost of mass redistribution. In computer vision, it serves as a robust alternative to nearest-neighbor assignments by finding structured correspondences between feature sets, with applications ranging from local matching~\cite{sarlin2020superglue} to object detection~\cite{ge2021ota}. Beyond matching, OT enables balanced clustering by enforcing marginal constraints that prevent samples from collapsing onto a few centers~\cite{caron2020unsupervised, wang2025pot}. This ensures that all prototypes are effectively utilized, maintaining the diversity required for compact representations. However, classical OT strictly enforces mass conservation, which can be undesirable in practical vision scenarios involving noisy features, redundant tokens, or partial correspondences. Unbalanced Optimal Transport (UOT)~\cite{frogner2015learning, liero2018optimal, chizat2018scaling} relaxes these marginal constraints by introducing soft penalties, allowing mass to be partially discarded or redistributed. This relaxation improves robustness in cases where exact matching is neither necessary nor semantically meaningful~\cite{ma2021learning, de2023unbalanced, lee2025unpaired}.

\begin{figure*}[!t]
    \centering
    \includegraphics[width=\linewidth]{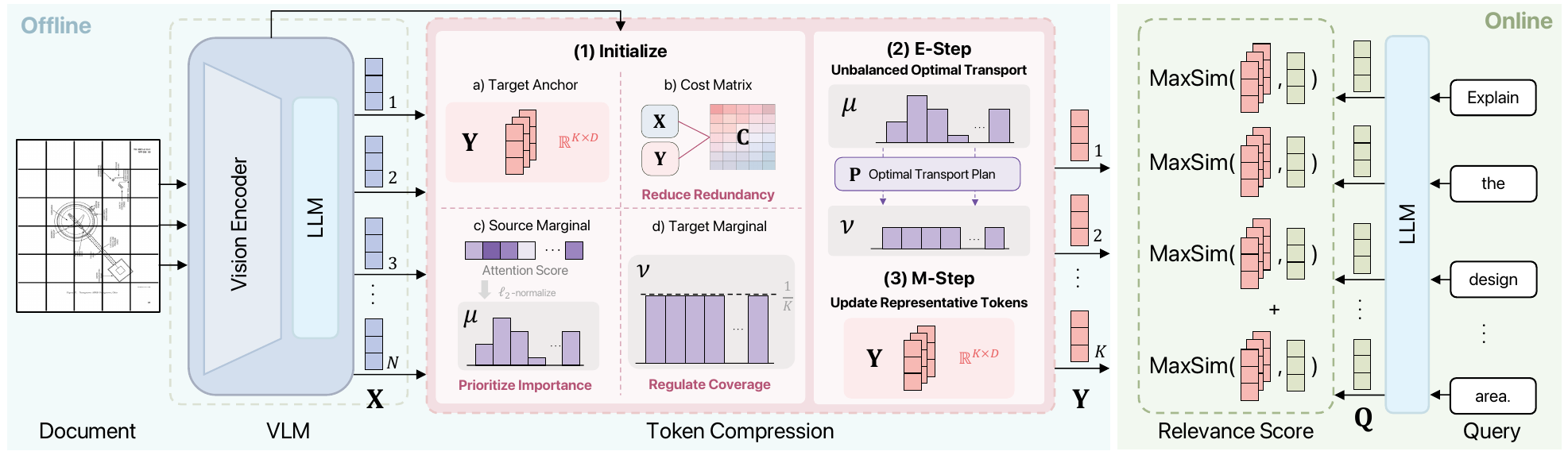}
    \caption{\textbf{Overview: Visual Token Compression.} Offline, $N$ visual tokens $\mathbf{X}$ are compressed into $K$ anchors $\mathbf{Y}$ ($K \ll N$). Our Unbalanced Optimal Transport (UOT) formulation jointly optimizes semantic redundancy (cost $\mathbf{C}$), token importance (source $\boldsymbol{\mu}$), and anchor capacity (target $\boldsymbol{\nu}$) to enable efficient online MaxSim retrieval.}
    \label{fig:overview}
\end{figure*}

\section{Proposed Method}
\label{sec:method}

\subsection{Problem Setup}
\label{subsec:problem_definition}
VLM-based Visual Document Retrieval (VDR) frameworks~\cite{faysse2025colpali} encode documents and queries into multi-vector representations. These models leverage the ViT architecture~\cite{dosovitskiy2021image} to represent a document image as $N$ visual tokens $\mathbf{X} = [\mathbf{x}_1, \dots, \mathbf{x}_N]^\top \in \mathbb{R}^{N \times D}$, where each $D$-dimensional token vector is $\ell_2$-normalized. Similarly, a text query is encoded into $M$ $\ell_2$-normalized tokens $\mathbf{Q} = [\mathbf{q}_1, \dots, \mathbf{q}_M]^\top \in \mathbb{R}^{M \times D}$. Late interaction~\cite{khattab2020colbert} then computes the relevance score via the MaxSim operation:
\begin{equation}
\label{eq:late_interaction}
s(\mathbf{Q}, \mathbf{X}) = \sum_{i=1}^M \max_{1 \le j \le N} \langle \mathbf{q}_i, \mathbf{x}_j \rangle.
\end{equation}
where $\langle\cdot,\cdot\rangle$ denotes the inner product. Due to the large number of tokens $N$ necessary for high-resolution images, this matching requires $\mathcal{O}(N M)$ operations per query-document pair, imposing severe computational and storage bottlenecks. To this end, token merging~\cite{bolya2023token, ma2025towards} compresses $\mathbf{X}$ into a compact set of $K$ representatives $\mathbf{Y} \in \mathbb{R}^{K \times D}$ ($K \ll N$) while preserving retrieval performance. We interpret this compression task as aligning dense patch-level features with a smaller set of representative anchors under the late-interaction objective. This perspective motivates a transport-based formulation, where token assignments are modeled as a mass redistribution problem.

\subsection{Preliminaries: Unbalanced Optimal Transport}
\label{subsec:preliminary}
Optimal Transport (OT) provides a principled framework for aligning discrete measures. Given source and target mass vectors $\boldsymbol{\mu} \in \mathbb{R}^N_+, \boldsymbol{\nu} \in \mathbb{R}^K_+$ and a cost matrix $\mathbf{C} \in \mathbb{R}^{N \times K}$, a transport plan $\mathbf{P} \in \mathbb{R}^{N \times K}_+$ maps mass from source to target elements to minimize the total cost. However, the strict mass conservation is often too restrictive when dealing with redundant or noisy data. To address this, Unbalanced Optimal Transport (UOT)~\cite{frogner2015learning, liero2018optimal} relaxes the marginal constraints via soft penalties:
\begin{equation}
\label{eq:unbalanced_ot}
\min_{\mathbf{P} \ge 0} \langle \mathbf{P}, \mathbf{C} \rangle_{F} - \gamma H(\mathbf{P}) + \rho \mathrm{KL}(\mathbf{P} \mathbf{1} \| \boldsymbol{\mu}) + \rho \mathrm{KL}(\mathbf{P}^\top \mathbf{1} \| \boldsymbol{\nu}),
\end{equation}
where $\langle\cdot,\cdot\rangle_{F}$ denotes the Frobenius inner product, $H(\mathbf{P})$ is the entropy regularization weighted by $\gamma > 0$, and $\rho > 0$ controls the degree of marginal relaxation. This formulation allows for flexible mass transport, which the generalized Sinkhorn algorithm~\cite{chizat2018scaling} optimizes efficiently.

\subsection{Token Compression via Unbalanced Optimal Transport}
\label{subsec:token_compression}
We formulate the compression of $N$ visual tokens $\mathbf{X}$ into $K$ representatives $\mathbf{Y}$ as a UOT-based mass redistribution problem. We address three objectives to ensure high-quality retrieval: minimizing semantic redundancy, preserving token importance, and regulating coverage capacity. While existing methods aggregate similar patches or prune less informative ones, they often neglect the coverage capacity of individual representatives. Unconstrained merging can collapse excessive patches into a single token, degrading the fine-grained localized features necessary for the MaxSim operation in~\eqref{eq:late_interaction}. Our framework unifies these objectives by mapping them to the transport cost and the relaxed marginal constraints, respectively.

\subsubsection{Initialization}
\label{subsubsec:initialization}
Let $\mathbf{Y} = [\mathbf{y}_1, \dots, \mathbf{y}_K]^\top \in \mathbb{R}^{K \times D}$ represent the $K$ tokens to optimize. We initialize these anchors, denoted as $\mathbf{Y}^{(0)}$, via a $k$-means++ strategy~\cite{arthur2007k}. We then apply $\ell_2$ normalization to these sampled anchors to project them onto the unit hypersphere. This seeding captures the visual distribution and accelerates convergence. We then define the transport cost and marginal distributions as follows:

\paragraph{Cost Matrix.} 
We penalize transport between dissimilar tokens to reduce semantic redundancy. Each entry $c_{ij}$ of $\mathbf{C} \in \mathbb{R}^{N \times K}$ is the cosine distance between the source token $\mathbf{x}_i$ and target anchor $\mathbf{y}_j$. Since these are $\ell_2$-normalized for all $i,j$, we have:
\begin{equation}
\label{eq:cost_matrix}
c_{ij} = 1 - \langle \mathbf{x}_i, \mathbf{y}_j \rangle.
\end{equation}
Minimizing this cost encourages semantically redundant patches to collapse into the same representative token.

\paragraph{Source Marginal.} 
To prioritize salient regions, we use the VLM’s self-attention weights as an importance proxy, since they naturally emphasize informative regions~\cite{djilali2023vision, faysse2025colpali}. Let $\mathbf{A} \in \mathbb{R}^{N \times N}$ be the attention matrix averaged over all transformer layers. The importance score $\tilde{a}_i$ is defined as the total incoming attention, $\tilde{a}_i = \sum_{n=1}^{N} a_{ni}$, where $a_{ni}$ denotes the element at the $n$-th row and $i$-th column. The $i$-th element $\mu_i$ of the source marginal distribution $\boldsymbol{\mu}$ is then given by:
\begin{equation}
\label{eq:source_mass}
\mu_i = \frac{\tilde{a}_i + \epsilon}{\sum_{n=1}^N (\tilde{a}_n + \epsilon)},
\end{equation}
where $\epsilon$ is a small constant for numerical stability.

\paragraph{Target Marginal.} To regulate the coverage of representative tokens, we define the target marginal $\boldsymbol{\nu} \in \mathbb{R}^K_+$ as a uniform distribution, where each element $\nu_j$ is given by:
\begin{equation}
\label{eq:target_mass}
\nu_j = \frac{1}{K}
\end{equation}
This establishes an equal capacity for each representative token, encouraging an equitable assignment of source mass. The KL penalty in~\eqref{eq:unbalanced_ot} relaxes strict capacity, allowing the plan to balance the uniform capacity prior with semantic density of document regions, maintaining localized features while avoiding rigid fixed marginals.

\begin{algorithm}[!t]
\caption{UOT-based Token Compression (PyTorch-style)}
\label{alg:token_compression}
\begin{lstlisting}[language=Python]
# X: Visual tokens (N, D)
# A: Attention matrix (N, N)
# K: Number of representative tokens
# T: Number of EM loop iterations

# 1. Initialize Marginal
a_tilde = A.sum(dim=0)
mu = l1_normalize(a_tilde, dim=0)
nu = l1_normalize(torch.ones(K), dim=0)

# 2. Initialize Anchors
Y = init_kmeans++(X, K)
Y = l2_normalize(Y, dim=-1)

# 3. Alternating Optimization (EM Loop)
for _ in range(T):
    # E-Step: Solve Unbalanced Optimal Transport
    C = 1.0 - (X @ Y.T)
    P = generalized_sinkhorn(C, mu, nu)
    
    # M-Step: Update Representative Tokens
    Y = P.T @ X
    Y = l2_normalize(Y, dim=-1)
\end{lstlisting}
\end{algorithm}

\begin{algorithm}[!t]
\caption{Generalized Sinkhorn Algorithm (PyTorch-style)}
\label{alg:generalized_sinkhorn}
\begin{lstlisting}[language=Python]
# C: Cost matrix (N, K)
# mu: Source marginal (N,), nu: Target marginal (K,)
# gamma: Entropy weight, rho: KL penalty weight
# L: Number of Sinkhorn iterations

log_mu, log_nu = mu.log(), nu.log()
log_u, log_v = torch.zeros_like(mu), torch.zeros_like(nu)

log_K = -C / gamma
fi = rho / (rho + gamma)

for _ in range(L):
    # Row Update
    t1 = log_K + log_v.unsqueeze(0)
    log_u = fi * (log_mu - torch.logsumexp(t1, dim=1))
    
    # Column Update
    t2 = log_K + log_u.unsqueeze(1)
    log_v = fi * (log_nu - torch.logsumexp(t2, dim=0))

return torch.exp(log_u.unsqueeze(1) + log_K + log_v.unsqueeze(0))
\end{lstlisting}
\end{algorithm}

\subsubsection{Alternating Optimization}
\label{subsubsec:alternating_optimization}
Since the representative tokens $\mathbf{Y}$ and their optimal assignments $\mathbf{P}$ are interdependent, we jointly solve for them using an Expectation-Maximization (EM)-style alternating optimization framework~\cite{dempster1977maximum}. The overall token compression procedure is summarized in~\cref{alg:token_compression}.

\paragraph{E-step.} At iteration $t$, we fix the current target anchors $\mathbf{Y}^{(t)}$ and compute the cost matrix $\mathbf{C}^{(t)}$ using~~\eqref{eq:cost_matrix}. Given the predefined marginals $\boldsymbol{\mu}$ and $\boldsymbol{\nu}$, we solve the UOT objective in~~\eqref{eq:unbalanced_ot} via the generalized Sinkhorn algorithm~\cite{chizat2018scaling}. To ensure numerical stability, we perform this optimization in the log domain, as detailed in~\cref{alg:generalized_sinkhorn}. This yields the optimal transport plan $\mathbf{P}^{(t)} \in \mathbb{R}^{N \times K}_+$. This soft assignment matrix balances three objectives: minimizing semantic distance, respecting attention-based source token importance, and adhering to the relaxed capacity limits of each cluster.

\paragraph{M-step.} Next, we fix the transport plan $\mathbf{P}^{(t)}$ and update the target anchors to minimize the expected transport cost. Given the cosine distance cost matrix, we update each anchor $j$ using a mass-weighted sum and $\ell_2$ normalization:
\begin{equation}
\label{eq:barycentric}
\mathbf{y}_j^{(t+1)} = \frac{\sum_{i=1}^N p_{ij}^{(t)} \mathbf{x}_i}{\left\| \sum_{i=1}^N p_{ij}^{(t)} \mathbf{x}_i \right\|_2}, \quad \forall j \in \{1, \dots, K\},
\end{equation}
where $p_{ij}^{(t)}$ is the $(i, j)$-th entry of $\mathbf{P}^{(t)}$. The transport plan $\mathbf{P}^{(t)}$ encodes both the attention-based source mass $\boldsymbol{\mu}$ and the relaxed uniform target capacity $\boldsymbol{\nu}$. Consequently, the updated tokens $\mathbf{Y}$ achieve a critical balance: the source mass ensures that informative tokens dominate the representations, while the regulated capacity prevents excessive aggregation into any single anchor. This dual effect preserves the localized feature diversity required for late interaction and effectively suppresses redundant tokens.

\section{Theoretical Analysis}
\label{sec:theoretical_analysis}
We theoretically analyze our UOT-based compression to ensure it preserves retrieval performance. Specifically, we bound the MaxSim approximation error, prove geometric robustness against worst-case queries, and demonstrate that these token-level bounds strictly preserve system-level document rankings. Detailed proofs for all theorems are provided in the supplementary material.

\subsection{MaxSim Approximation Error Bound}
\label{subsec:maxsim_bound}
We define the retrieval error as the difference between the original MaxSim score $s(\mathbf{Q}, \mathbf{X})$ and the compressed score $s(\mathbf{Q}, \mathbf{Y})$ defined in~~\eqref{eq:late_interaction}. The following theorem establishes an upper bound on this error using a transport plan $\mathbf{P}$.

\begin{theorem}[MaxSim Approximation Error Bound]
\label{thm:maxsim_bound}
Assume all token vectors in $\mathbf{Q}, \mathbf{X}, \mathbf{Y}$ are $\ell_2$-normalized. Let $\mathbf{C} \in \mathbb{R}^{N \times K}$ be the cosine-distance cost matrix where $c_{ij} = 1 - \langle \mathbf{x}_i, \mathbf{y}_j \rangle$. For a transport plan $\mathbf{P} \in \mathbb{R}^{N \times K}_+$, let $i_m^\star = \arg\max_{1 \le i \le N} \langle \mathbf{q}_m, \mathbf{x}_i \rangle$ be the best-matching original token for query token $\mathbf{q}_m$. If the row mass $r_m = \sum_{j=1}^K p_{i_m^\star j}$ is positive, the MaxSim degradation is bounded by:
\begin{equation}
\label{eq:maxsim_bound_main}
s(\mathbf{Q}, \mathbf{X}) - s(\mathbf{Q}, \mathbf{Y}) \le \sqrt{2} \sum_{m=1}^{M} \sqrt{ \frac{1}{r_m} \sum_{j=1}^{K} p_{i_m^\star j} c_{i_m^\star j} }.
\end{equation}
\end{theorem}

\subsubsection{Remark.} 
This bound justifies our formulation. The term $\sum_j p_{i_m^\star j} c_{i_m^\star j}$ measures the transport cost for tokens that dominate the MaxSim score. Minimizing $\langle \mathbf{P}, \mathbf{C} \rangle$ therefore reduces the approximation error in informative regions. Additionally, our importance-weighted marginal $\boldsymbol{\mu}$ assigns more mass to these likely maximizers, which further tightens the bound.

\subsection{Worst-Case Query Robustness}
\label{subsec:worst_case}
We analyze the robustness of our framework against any possible unit-norm query. Let $\Delta(\mathbf{X}, \mathbf{Y})$ denote the worst-case MaxSim gap for a single query token $\mathbf{q}$. The following theorem connects this retrieval gap to the geometric covering radius between the original tokens and their representatives.

\begin{theorem}[Worst-case Query Robustness]
\label{thm:worst_case}
Assume all token vectors are $\ell_2$-normalized, and let $c_{ij} = 1 - \langle \mathbf{x}_i, \mathbf{y}_j \rangle$ be the cosine cost. The worst-case MaxSim gap is bounded by the geometric covering radius:
\begin{equation}
\label{eq:hausdorff_main}
\Delta(\mathbf{X}, \mathbf{Y}) := \sup_{\|\mathbf{q}\|_2=1} \left( \max_{i} \langle \mathbf{q}, \mathbf{x}_i \rangle - \max_{j} \langle \mathbf{q}, \mathbf{y}_j \rangle \right) \le \max_{i} \min_{j} \|\mathbf{x}_i - \mathbf{y}_j\|_2.
\end{equation}
Furthermore, for any transport plan $\mathbf{P} \in \mathbb{R}^{N \times K}_+$ with row mass $r_i = \sum_{j=1}^K p_{ij} > 0$, this gap is upper-bounded by the transport cost:
\begin{equation}
\label{eq:worst_case_transport}
\Delta(\mathbf{X}, \mathbf{Y}) \le \sqrt{2} \max_i \sqrt{ \frac{1}{r_i} \sum_{j=1}^K p_{ij} c_{ij} }.
\end{equation}
\end{theorem}

\subsubsection{Remark.}
This bound links the worst-case retrieval error to the geometric covering radius. While unconstrained compression may discard minor visual details and expand this radius, \cref{thm:worst_case} proves that minimizing the UOT cost strictly bounds it. By ensuring every original token $\mathbf{x}_i$ remains geometrically close to a representative, our method prevents search blind spots, preserving fine-grained features necessary for robust long-tail retrieval.

\subsection{Ranking Stability under Compression}
\label{subsec:ranking_stability}
We show that bounding the per-document MaxSim degradation preserves retrieval rankings. For a query $\mathbf{Q}$ and a document collection $\mathcal{D}$, let $s_d$ and $\tilde{s}_d$ denote the original and compressed scores for document $d$, respectively. Assume a deviation bound $|s_d - \tilde{s}_d| \le \varepsilon_d(\mathbf{Q})$. As established, the UOT transport cost directly governs this bound $\varepsilon_d(\mathbf{Q})$.

\begin{theorem}[Ranking and Top-$k$ Stability]
\label{thm:ranking_stability}
Assume the deviation bound $|s_d - \tilde{s}_d| \le \varepsilon_d(\mathbf{Q})$ holds.
\begin{enumerate}
\item \textbf{Pairwise Stability:} For any two documents $a$ and $b$, if their original score gap satisfies $s_a - s_b > \varepsilon_a(\mathbf{Q}) + \varepsilon_b(\mathbf{Q})$, their relative ranking is preserved after compression ($\tilde{s}_a > \tilde{s}_b$).
\item \textbf{Top-$k$ Stability:} Let $\mathcal{T}_k$ be the set of the top-$k$ documents under the original scores. Define the top-$k$ gap as $\Delta_k(\mathbf{Q}) = s_{(k)} - s_{(k+1)}$, where $s_{(k)}$ is the $k$-th highest original score. If $\Delta_k(\mathbf{Q}) > \max_{d \in \mathcal{T}_k} \varepsilon_d(\mathbf{Q}) + \max_{d \notin \mathcal{T}_k} \varepsilon_d(\mathbf{Q})$, the top-$k$ set $\mathcal{T}_k$ is exactly preserved under the compressed scores.
\end{enumerate}
\end{theorem}

\subsubsection{Remark.} 
\Cref{thm:ranking_stability} bridges our token-level transport bounds with the system-level retrieval objective. Specifically, realizing the deviation $\varepsilon_d(\mathbf{Q})$ via \cref{thm:maxsim_bound} yields an explicit sufficient condition for ranking preservation based entirely on transport costs. Therefore, minimizing the UOT cost  maximizes the theoretical guarantees of ranking stability.

\section{Experiments}
\label{sec:experiment}

\subsection{Experimental Setup}
\subsubsection{Dataset.}
We evaluate our framework on ViDoRe V1~\cite{faysse2025colpali} and V3~\cite{loison2026vidorev3}, standard benchmarks for visually rich document retrieval. ViDoRe V1 contains multiple subtasks covering diverse document types, languages, and real-world scenarios. We report results on the standard academic and practical tasks. ViDoRe V3 contains domain-specific datasets with graded relevance annotations, allowing multiple relevant pages per query instead of V1's single-positive setup.

\paragraph{ViDoRe V1 Academic Tasks.}
\label{subsubsec:academic_tasks.}
This split adapts VQA benchmarks by treating questions as queries and pages as documents. InfoVQA (InfoQ)~\cite{mathew2022infographicvqa} features dense infographics, while arXivQA (ArxivQ)~\cite{li2024multimodal} focuses on scientific figures and trends. TabFQuAD (TabF) provides French table queries, and TAT-DQA (TATQ)~\cite{zhu2022towards} mixes tables with text. DocVQA (DocQ)~\cite{mathew2021docvqa} covers real-world free-form documents (\eg, receipts) requiring key-field localization.

\paragraph{ViDoRe V1 Practical Tasks.}
\label{subsubsec:practical_tasks.}
This split is designed to evaluate retrieval in realistic settings. It includes five domain-specific subsets: Energy, Government (Gov.), Healthcare (Health.), Artificial Intelligence (AI.), and French environmental reports (Shift.). Each subset contains 1,000 document pages and 100 queries. The topic corpora are intentionally narrow to increase similarity among candidate pages, making the ranking problem more challenging and representative of real-world retrieval scenarios.

\paragraph{ViDoRe V3 English-Only Tasks.}
\label{subsubsec:vidore_v3.}
This benchmark contains 26K pages and 3.1K queries translated into 6 languages. We evaluate on ViDoRe V3 using English queries only, covering five domain-specific subsets: Computer Science, Finance, HR report, Industrial, and Pharmaceuticals. These splits contain 13K document pages and 1.5K queries in a domain-focused retrieval setting. The benchmark includes diverse query types ranging from simple fact-seeking questions to queries requiring structural or semantic understanding of documents. Due to high similarity among candidate pages and the increased query complexity, the retrieval task becomes more challenging. To keep the main evaluation focused on domain-specific retrieval, we use English queries in the main experiments and report results on the remaining language queries in the supplementary material.

\subsubsection{Evaluation Metrics.}
We evaluate retrieval performance using nDCG@5 and Recall@$1$. Specifically, nDCG@5 measures ranked relevance within the top-$5$ results, and Recall@1 evaluates whether the ground-truth page is ranked first. Because ViDoRe V1 uses binary single-relevance labels while V3 provides multiple graded labels, we ensure consistent evaluation by adopting the capped Recall@$k$ from BEIR~\cite{thakur2021beir}. This metric computes the fraction of relevant documents retrieved within the top-$k$, capped by the total relevant documents per query. For efficiency, we report the merging factor (MF), the ratio of original visual tokens $N$ to representative tokens $K$, which directly reflects memory and scoring cost reductions. Also, we measure average query latency as the mean time to compute query-document relevance scores.

\subsubsection{Implementation Details}
We evaluate our framework using Qwen2-VL-2B (Qwen2)~\cite{wang2024qwen2} and PaliGemma-3B (Pali)~\cite{beyer2024paligemma} for ViDoRe V1, and Qwen2/Qwen2.5-VL-3B (Qwen2.5) for ViDoRe V3. For a fair comparison, all compression methods are evaluated under the merging factors (MF) $\in \{5, 10, 25, 50\}$. In \cref{alg:token_compression,alg:generalized_sinkhorn}, we use $T=20$ EM iterations, $L=3$ Sinkhorn iterations, and $\rho=5$ for all models. The entropic regularization $\gamma$ is set to $0.04$ for Qwen2/Qwen2.5 and $0.01$ for Pali. All experiments are conducted on a single NVIDIA A6000 GPU.

\begin{table*}[!t]
    \centering
    \tiny
    \renewcommand{\arraystretch}{0.935}
    \caption{\textbf{Results on ViDoRe V1 academic tasks.} nDCG@5/Recall@1 ($\uparrow$) and Time (ms/q). Best compression results per MF are bolded.}
    \vspace{-0.2cm}
    {\setlength{\tabcolsep}{3pt}
    \begin{tabular}{L{0.155\linewidth}|C{0.035\linewidth}|C{0.060\linewidth}|*{5}{C{0.0975\linewidth}|}C{0.0975\linewidth}}
        \toprule
        \textbf{Method} & 
        \textbf{MF} &
        \textbf{Time} &
        \textbf{ArxivQ} &
        \textbf{DocQ} &
        \textbf{InfoQ} &
        \textbf{TabF} &
        \textbf{TATQ} &
        \textbf{Avg.}\\
        \midrule
        \rowcolor{lightgray!35}
        \multicolumn{9}{l}{\textit{Base model: Qwen2-VL-2B}}\\
        \midrule
        ColQwen2 & - & 0.63 & 86.8/80.4 & 56.0/48.0 & 90.6/86.2 & 88.7/81.4 & 78.5/66.0 & 80.1/72.4 \\
        \midrule
        ColQwen2+Prune & \multirow{3}{*}{5} & \multirow{3}{*}{0.16} & 83.6/76.8 & 54.7/46.4 & 86.7/80.2 & 85.8/78.2 & 71.9/59.2 & 76.5/68.2 \\
        Light-ColQwen2& &
        & 85.8/79.0 & 54.5/46.2 & 87.4/81.6 & 87.3/79.6 & 75.3/61.8 & 78.1/69.6\\
        Ours &  &
        & \textbf{86.3}/\textbf{79.6} & \textbf{56.4}/\textbf{48.4} & \textbf{89.0}/\textbf{84.2} & \textbf{89.5}/\textbf{83.2} & \textbf{77.0}/\textbf{64.0} & \textbf{79.6}/\textbf{71.9} \\
        \midrule
        ColQwen2+Prune & \multirow{3}{*}{10} & \multirow{3}{*}{0.11} & 80.8/74.6 & 51.1/42.4 & 80.2/72.2 & 82.3/73.6 & 64.9/51.9 & 71.9/62.9 \\
        Light-ColQwen2 &  & 
        & 84.9/78.6 & 53.8/\textbf{45.6} & 82.4/74.2 & 87.3/79.3 & 70.6/57.4 & 75.8/67.0\\
        Ours &  &
        & \textbf{85.8}/\textbf{79.2} & \textbf{54.1}/44.6 & \textbf{88.5}/\textbf{83.2} & \textbf{90.0}/\textbf{84.6} & \textbf{74.6}/\textbf{61.3} & \textbf{78.6}/\textbf{70.6} \\
        \midrule
        ColQwen2+Prune & \multirow{3}{*}{25} &\multirow{3}{*}{0.08} &75.2/68.6 & 42.2/34.2 & 72.3/62.2 & 76.8/67.9 & 53.1/40.4 & 63.9/54.7 \\
        Light-ColQwen2 & & 
        & 81.8/75.6 & 48.7/39.4 & 72.9/63.0 & 86.5/78.6 & 56.5/45.1 & 69.3/60.3\\
        Ours &  &
        & \textbf{83.5}/\textbf{76.6} & \textbf{50.8}/\textbf{41.4} & \textbf{85.8}/\textbf{79.2} & \textbf{87.7}/\textbf{81.4} & \textbf{69.0}/\textbf{55.5} & \textbf{75.4}/\textbf{66.8} \\
        \midrule
        ColQwen2+Prune & \multirow{3}{*}{50} &\multirow{3}{*}{0.07} & 66.3/58.6 & 35.8/28.6 & 65.7/54.4 & 65.0/54.3 & 43.1/31.9 & 55.2/45.6 \\
        Light-ColQwen2 &  & 
        & 77.4/69.2 & 39.4/30.4 & 65.8/55.8 & 81.1/71.8 & 44.5/34.4 & 61.6/52.3\\
        Ours &  &
        & \textbf{82.2}/\textbf{75.4} & \textbf{45.2}/\textbf{36.4} & \textbf{81.2}/\textbf{73.6} & \textbf{86.8}/\textbf{79.6} & \textbf{60.1}/\textbf{47.1} & \textbf{71.3}/\textbf{62.4} \\
        \midrule

        \rowcolor{lightgray!35}
        \multicolumn{9}{l}{\textit{Base model: PaliGemma-3B}}\\
        \midrule
        ColPali & - & 0.80
        & 77.6/71.0 & 49.6/41.6 & 79.1/72.2 & 85.3/77.9 & 65.2/52.8 & 71.4/63.1 \\
        \midrule
        ColPali+Prune & \multirow{3}{*}{5} &\multirow{3}{*}{ 0.20} & 76.0/70.0 & 47.9/39.2 & 76.1/68.2 & \textbf{85.0}/\textbf{77.9} & 61.3/48.6 & 69.3/60.8 \\
        Light-ColPali &  & 
        & 75.8/\textbf{70.2} & 42.3/34.8 & 76.5/68.8 & 83.9/76.1 & 57.4/43.7 & 67.2/58.7\\
        Ours &  &
        & \textbf{76.4}/\textbf{70.2} & \textbf{49.2}/\textbf{41.6} & \textbf{77.9}/\textbf{71.6} & 80.5/72.5 & \textbf{63.9}/\textbf{51.9} & \textbf{69.6}/\textbf{61.6} \\
        \midrule
        ColPali+Prune & \multirow{3}{*}{10} &\multirow{3}{*}{0.13} & 72.9/66.4 & 45.0/36.6 & 72.8/63.4 & 82.2/72.1 & 54.0/41.7 & 65.4/56.0 \\
        Light-ColPali &  & 
        & 72.9/66.4 & 35.0/27.0 & 59.3/51.2 & 80.8/71.4 & 41.1/28.7 & 57.8/48.9\\
        Ours &  &
        & \textbf{75.7}/\textbf{69.8} & \textbf{48.1}/\textbf{40.2} & \textbf{77.4}/\textbf{70.8} & \textbf{81.3}/\textbf{74.6} & \textbf{62.7}/\textbf{49.5} & \textbf{69.0}/\textbf{61.0} \\
        \midrule
        ColPali+Prune & \multirow{3}{*}{25} &\multirow{3}{*}{0.08} &67.4/59.4 & 37.4/29.6 & 64.6/56.4 & 71.4/60.7 & 43.7/31.8 & 56.9/47.6 \\
        Light-ColPali &  & 
        & 63.1/53.8 & 25.2/19.2 & 58.6/47.8 & 71.4/60.7 & 27.3/17.3 & 49.1/39.8\\
        Ours &  &
        & \textbf{75.0}/\textbf{68.2} & \textbf{45.4}/\textbf{39.0} & \textbf{76.3}/\textbf{69.6} & \textbf{78.7}/\textbf{71.1} & \textbf{58.0}/\textbf{45.3} & \textbf{66.7}/\textbf{58.6} \\
        \midrule
        ColPali+Prune & \multirow{3}{*}{50} &\multirow{3}{*}{0.07} & 60.1/51.0 & 29.3/23.0 & 57.7/49.6 & 60.9/49.6 & 33.0/24.1 & 48.2/39.5 \\
        Light-ColPali & & 
        & 51.0/43.4 & 21.8/16.8 & 45.1/35.4 & 62.9/51.1 & 21.0/12.2 & 40.4/31.8 \\
        Ours &  &
        & \textbf{73.8}/\textbf{67.4} & \textbf{42.8}/\textbf{37.2} & \textbf{73.6}/\textbf{65.6} & \textbf{76.5}/\textbf{68.6} & \textbf{51.9}/\textbf{38.4} & \textbf{63.7}/\textbf{55.4} \\
        \bottomrule
    \end{tabular}}
    \label{tab:main_academic}
\end{table*}

\begin{table*}[!t]
    \centering
    \tiny
    \renewcommand{\arraystretch}{0.935}
    \caption{\textbf{Results on ViDoRe V1 practical tasks.} nDCG@5/Recall@1 ($\uparrow$) and Time (ms/q). Best compression results per MF are bolded.}
    \vspace{-0.2cm}
    {\setlength{\tabcolsep}{3pt}
    \begin{tabular}{L{0.155\linewidth}|C{0.035\linewidth}|C{0.060\linewidth}|*{5}{C{0.0975\linewidth}|}C{0.0975\linewidth}}
        \toprule
        \textbf{Method} & 
        \textbf{MF} & 
        \textbf{Time} & 
        \textbf{Shift.} & 
        \textbf{AI.} & 
        \textbf{Energy.} & 
        \textbf{Gov.} & 
        \textbf{Health.} & 
        \textbf{Avg.}\\
        \midrule

        \rowcolor{lightgray!35}
        \multicolumn{9}{l}{\textit{Base model: Qwen2-VL-2B}}\\
        \midrule
        ColQwen2 & - &1.06& 87.5/77.0 & 86.7/67.0 & 85.6/67.0 & 84.3/63.0 & 85.2/64.0 & 85.9/67.6 \\
        \midrule
        ColQwen2+Prune & \multirow{3}{*}{5} & \multirow{3}{*}{0.27} & 78.1/60.0 & 85.2/64.0 & \textbf{85.7}/\textbf{69.0} & \textbf{85.9}/\textbf{67.0} & 82.7/60.0 & 83.5/64.0 \\
        Light-ColQwen2 &  &
        & 81.7/67.0 & \textbf{85.6}/\textbf{65.0} & 83.3/64.0 & 82.7/61.0 & \textbf{85.2}/\textbf{63.0} & 83.7/64.0\\
        Ours &  &
        & \textbf{86.9}/\textbf{74.0} & 84.7/62.0 & 83.2/62.0 & 85.6/\textbf{67.0} & 84.2/61.0 & \textbf{84.9}/\textbf{65.2} \\
        \midrule
        ColQwen2+Prune & \multirow{3}{*}{10} & \multirow{3}{*}{0.19} & 70.0/49.0 & 82.7/61.0 & 82.2/\textbf{66.0} & 84.8/65.0 & 81.9/58.0 & 80.3/59.8 \\
        Light-ColQwen2 & & 
        & 75.1/58.0 & 83.4/62.0 & 79.1/59.0 & 80.5/57.0 & 83.5/\textbf{62.0} & 80.3/59.6\\
        Ours &  &
        & \textbf{83.5}/\textbf{66.0} & \textbf{84.5}/\textbf{63.0} & \textbf{83.7}/64.0 & \textbf{86.1}/\textbf{67.0} & \textbf{83.9}/\textbf{62.0} & \textbf{84.3}/\textbf{64.4} \\
        \midrule
        ColQwen2+Prune & \multirow{3}{*}{25} & \multirow{3}{*}{0.15} &54.9/37.0 & 73.7/51.0 & 73.0/51.0 & 75.8/55.0 & 79.4/54.0 & 71.4/49.6 \\
        Light-ColQwen2 &  & 
        & 58.1/42.0 & 76.7/57.0 & 73.7/50.0 & 78.1/54.0 & 74.1/52.0 & 72.1/51.0\\
        Ours &  &
        & \textbf{81.5}/\textbf{70.0} & \textbf{83.4}/\textbf{63.0} & \textbf{83.3}/\textbf{68.0} & \textbf{82.7}/\textbf{61.0} & \textbf{81.6}/\textbf{58.0} & \textbf{82.5}/\textbf{64.0} \\
        \midrule
        ColQwen2+Prune & \multirow{3}{*}{50} & \multirow{3}{*}{0.13} &41.4/29.0 & 66.3/46.0 & 65.1/48.0 & 70.2/49.0 & 74.9/50.0 & 63.6/44.4 \\
        Light-ColQwen2 &  &
        & 38.0/23.0 & 64.6/42.0 & 64.0/43.0 & 69.2/43.0 & 65.5/44.0 & 60.3/39.0\\
        Ours &  &
        & \textbf{72.2}/\textbf{57.0} & \textbf{84.0}/\textbf{66.0} & \textbf{77.9}/\textbf{58.0} & \textbf{82.4}/\textbf{62.0} & \textbf{82.2}/\textbf{60.0} & \textbf{79.7}/\textbf{60.6} \\
        \midrule

        \rowcolor{lightgray!35}
        \multicolumn{9}{l}{\textit{Base model: PaliGemma-3B}}\\
        \midrule
        ColPali & - &0.80
        & 74.6/58.0 & 85.2/66.0 & 85.8/69.0 & 85.9/67.0 & 81.6/57.0 & 82.6/63.4 \\
        \midrule
        ColPali+Prune & \multirow{3}{*}{5} & \multirow{3}{*}{0.32} &68.5/54.0 & 81.6/\textbf{61.0} & 84.2/64.0 & 84.5/64.0 & 80.3/58.0 & 79.8/60.2 \\
        Light-ColPali &  & 
        & 58.6/42.0 & \textbf{83.2}/60.0 & 81.1/62.0 & 76.3/54.0 & 75.1/50.0 & 74.9/53.6\\
        Ours &  &
        & \textbf{77.0}/\textbf{62.0} & 83.1/\textbf{61.0} & \textbf{85.3}/\textbf{70.0} & \textbf{86.2}/\textbf{66.0} & \textbf{81.4}/\textbf{59.0} & \textbf{82.6}/\textbf{63.6} \\
        \midrule
        ColPali+Prune & \multirow{3}{*}{10} & \multirow{3}{*}{0.22} & 59.1/50.0 & 80.6/58.0 & 77.5/56.0 & 78.6/57.0 & 78.7/56.0 & 74.9/55.4 \\
        Light-ColPali &  & 
        & 37.4/27.0 & 66.4/42.0 & 67.8/46.0 & 63.9/41.0 & 66.3/43.0 & 60.4/39.8\\
        Ours &  &
        & \textbf{75.2}/\textbf{64.0} & \textbf{84.7}/\textbf{67.0} & \textbf{83.2}/\textbf{63.0} & \textbf{84.1}/\textbf{63.0} & \textbf{82.2}/\textbf{61.0} & \textbf{81.9}/\textbf{63.6} \\
        \midrule
        ColPali+Prune & \multirow{3}{*}{25} & \multirow{3}{*}{0.15} & 46.4/37.0 & 73.3/51.0 & 63.5/46.0 & 69.9/49.0 & 68.3/43.0 & 64.3/45.2 \\
        Light-ColPali &  &
        & 25.1/15.0 & 54.8/33.0 & 51.1/31.0 & 51.6/33.0 & 53.8/29.0 & 47.3/28.2\\
        Ours &  &
        & \textbf{69.8}/\textbf{57.0} & \textbf{82.7}/\textbf{62.0} & \textbf{85.2}/\textbf{71.0} & \textbf{84.1}/\textbf{63.0} & \textbf{82.3}/\textbf{63.0} & \textbf{80.8}/\textbf{63.2} \\
        \midrule
        ColPali+Prune & \multirow{3}{*}{50} & \multirow{3}{*}{0.14} & 32.9/24.0 & 62.2/39.0 & 51.8/38.0 & 55.0/36.0 & 56.2/35.0 & 51.6/34.4 \\
        Light-ColPali &  & 
        & 21.1/10.0 & 50.0/25.0 & 46.1/27.0 & 40.4/24.0 & 50.0/30.0 & 41.5/23.2\\
        Ours &  &
        & \textbf{62.8}/\textbf{44.0} & \textbf{79.9}/\textbf{59.0} & \textbf{81.1}/\textbf{63.0} & \textbf{83.7}/\textbf{64.0} & \textbf{79.9}/\textbf{59.0} & \textbf{77.5}/\textbf{57.8} \\
        \bottomrule
    \end{tabular}}
    \label{tab:main_practical}
\end{table*}

\begin{table*}[!t]
    \centering
    \tiny 
    \renewcommand{\arraystretch}{0.935}
    \caption{\textbf{Results on ViDoRe V3 english-only tasks.} nDCG@5/Recall@1 ($\uparrow$) and Time (ms/q). Best compression results per MF are bolded.}
    \vspace{-0.2cm}
    {\setlength{\tabcolsep}{2.8pt}
    \begin{tabular}{L{0.169\linewidth}|C{0.034\linewidth}|C{0.057\linewidth}|*{5}{C{0.0965\linewidth}|}C{0.0965\linewidth}}
        \toprule
        \textbf{Method} & 
        \textbf{MF} & 
        \textbf{Time} & 
        \textbf{C.S.} & 
        \textbf{Fin.} & 
        \textbf{H.R.} & 
        \textbf{Ind.} & 
        \textbf{Phar.} & 
        \textbf{Avg.}\\
        \midrule

        \rowcolor{lightgray!35}
        \multicolumn{9}{l}{\textit{Base model: Qwen2-VL-2B}}\\
        \midrule
        ColQwen2 & - & 2.50 & 67.7/75.0 & 54.5/54.5 & 55.4/55.4 & 49.3/53.2 & 58.8/59.1 & 57.1/59.4 \\
        \midrule
        ColQwen2+Prune & \multirow{3}{*}{5} & \multirow{3}{*}{0.63} & \textbf{67.7}/75.8 & \textbf{52.7}/53.7 & 51.5/52.2 & 44.7/45.9 & 55.9/56.6 & 54.5/56.8 \\
        
        Light-ColQwen2 &  & 
        & 67.3/\textbf{76.4} & 48.8/50.0 & 48.8/47.5 & 45.9/\textbf{48.6} & 56.4/55.8 & 53.4/55.7 \\
        Ours &  &
        & 67.5/75.5 & 52.3/\textbf{55.2} & \textbf{54.0}/\textbf{55.0} & \textbf{46.7}/48.2 & \textbf{57.6}/\textbf{58.2} & \textbf{55.6}/\textbf{58.4} \\
        \midrule
        ColQwen2+Prune & \multirow{3}{*}{10} & \multirow{3}{*}{0.41} & 62.5/67.9 & 46.3/48.5 & 48.1/\textbf{51.9} & 37.7/38.9 & 52.1/50.8 & 49.3/51.6 \\

        Light-ColQwen2 &  & 
        & 64.8/73.6 & 46.3/48.7 & 45.4/45.9 & 41.8/42.6 & 55.4/\textbf{54.9} & 50.7/53.1 \\
        Ours &  &
        & \textbf{67.2}/\textbf{76.4} & \textbf{48.8}/\textbf{51.0} & \textbf{53.1}/50.9 & \textbf{45.9}/\textbf{48.6} & \textbf{56.0}/54.7 & \textbf{54.2}/\textbf{56.3} \\
        \midrule
        ColQwen2+Prune & \multirow{3}{*}{25} & \multirow{3}{*}{0.29} & 56.7/60.5 & 35.8/36.9 & 39.5/38.4 & 32.0/31.1 & 43.1/43.1 & 41.4/42.0 \\

        Light-ColQwen2 & & 
        & 58.7/66.2 & 34.8/38.4 & 35.0/36.8 & 35.4/35.6 & 49.8/48.9 & 42.7/45.2 \\
        Ours &  &
        & \textbf{64.6}/\textbf{71.8} & \textbf{43.5}/\textbf{45.8} & \textbf{48.5}/\textbf{47.2} & \textbf{40.0}/\textbf{42.6} & \textbf{53.0}/\textbf{52.2} & \textbf{49.9}/\textbf{51.9} \\
        \midrule
        ColQwen2+Prune & \multirow{3}{*}{50} & \multirow{3}{*}{0.26} & 51.0/57.7 & 27.7/29.4 & 32.6/33.3 & 27.4/27.2 & 35.7/34.3 & 34.9/36.4 \\

        Light-ColQwen2 &  &
        & 49.0/56.5 & 19.2/21.0 & 24.2/25.5 & 27.2/25.7 & 38.4/38.2 & 31.6/33.4 \\
        Ours &  &
        & \textbf{60.9}/\textbf{69.0} & \textbf{38.9}/\textbf{41.0} & \textbf{45.8}/\textbf{47.5} & \textbf{35.0}/\textbf{35.6} & \textbf{50.5}/\textbf{51.4} & \textbf{46.2}/\textbf{48.9} \\
        \midrule

        \rowcolor{lightgray!35}
        \multicolumn{9}{l}{\textit{Base model: Qwen2.5-VL-3B}}\\
        \midrule
        ColQwen2.5 & - & 2.39
        & 69.8/75.5 & 52.7/56.8 & 55.8/56.9 & 48.9/51.1 & 60.6/61.8 & 57.6/60.4 \\
        \midrule
        ColQwen2.5+Prune & \multirow{3}{*}{5} & \multirow{3}{*}{0.60} & 66.0/71.6 & 48.7/54.4 & 51.9/52.8 & 41.8/42.4 & 57.0/58.8 & 53.1/56.0 \\

        Light-ColQwen2.5 &  &
        & 63.5/69.9 & 43.5/44.8 & 44.5/46.2 & 42.0/45.4 & 56.3/56.0 & 48.4/51.6 \\
        Ours &  &
        & \textbf{69.3}/\textbf{77.3} & \textbf{52.2}/\textbf{57.4} & \textbf{55.7}/\textbf{55.3} & \textbf{48.0}/\textbf{50.4} & \textbf{59.7}/\textbf{60.4} & \textbf{57.0}/\textbf{60.2} \\
        \midrule
        ColQwen2.5+Prune & \multirow{3}{*}{10} & \multirow{3}{*}{0.41} & 62.8/68.4 & 39.5/44.0 & 45.8/46.9 & 37.4/36.0 & 53.4/56.0 & 47.8/50.3 \\

        Light-ColQwen2.5 & & 
        & 50.5/55.1 & 32.3/34.5 & 31.4/29.9 & 32.7/28.9 & 51.9/52.5 & 39.8/40.2 \\
        Ours &  &
        & \textbf{69.7}/\textbf{74.1} & \textbf{50.8}/\textbf{56.1} & \textbf{53.5}/\textbf{53.1} & \textbf{45.8}/\textbf{47.5} & \textbf{60.1}/\textbf{62.9} & \textbf{56.0}/\textbf{58.7} \\
        \midrule
        ColQwen2.5+Prune & \multirow{3}{*}{25} & \multirow{3}{*}{0.30} & 55.4/62.8 & 34.2/35.6 & 36.1/36.5 & 29.3/27.9 & 46.2/46.2 & 40.2/41.8 \\

        Light-ColQwen2.5 &  &
        & 44.6/48.1 & 23.2/21.0 & 20.1/17.3 & 21.6/21.1 & 38.6/36.8 & 29.6/28.9 \\
        Ours &  &
        & \textbf{69.4}/\textbf{78.2} & \textbf{46.1}/\textbf{50.3} & \textbf{51.2}/\textbf{53.8} & \textbf{41.5}/\textbf{37.0} & \textbf{56.7}/\textbf{56.9} & \textbf{53.0}/\textbf{55.2} \\
        \midrule
        ColQwen2.5+Prune & \multirow{3}{*}{50} & \multirow{3}{*}{0.26} & 51.4/60.0 & 24.9/23.9 & 26.1/28.6 & 19.0/18.0 & 38.3/37.4 & 31.9/33.6 \\

        Light-ColQwen2.5 &  & 
        & 45.2/50.5 & 19.6/18.4 & 21.6/22.0 & 13.6/11.6 & 31.0/25.5 & 26.2/25.6 \\
        Ours &  &
        & \textbf{66.4}/\textbf{73.1} & \textbf{40.6}/\textbf{45.2} & \textbf{47.3}/\textbf{50.3} & \textbf{37.8}/\textbf{36.6} & \textbf{54.3}/\textbf{55.2} & \textbf{49.3}/\textbf{52.1} \\
        \bottomrule
    \end{tabular}}
    \label{tab:main_vidorev3}
\end{table*}

\subsection{Main Results}
\label{subsec:main_results}

\subsubsection{Quantitative Comparison.}
\label{subsubsec:quantitative_comp}
We set the uncompressed multi-vector retrievers~\cite{faysse2025colpali} (ColQwen2, ColPali, ColQwen2.5) as upper bounds. We compare our method against two training-free compression baselines: (i) attention-based pruning~\cite{faysse2025colpali} (e.g., ColQwen2+Prune), which discards low-attention tokens, and (ii) distance-based clustering~\cite{ma2025towards} (e.g., Light-ColQwen2), which merges redundant tokens via hierarchical clustering. For a fair comparison, all methods utilize the same frozen VLM backbones and operate under identical token budgets (MF $\in \{5, 10, 25, 50\}$).

\Cref{tab:main_academic,tab:main_practical,tab:main_vidorev3} summarize the results across the ViDoRe V1 and V3 benchmarks. While all methods perform comparably under lenient budgets (MF=5), heuristic baselines degrade severely at extreme compression (MF=50), as unconstrained clustering forcibly merges disparate visual elements and induces representation collapse. In contrast, our framework remains highly robust by explicitly regulating token capacity and adaptively redistributing mass to preserve fine-grained localized features. Consequently, our method achieves substantial gains over Light-ColQwen baselines at MF=50, outperforming them by +19.4 nDCG@5 on the V1 Practical tasks (Qwen2) and +23.1 nDCG@5 on the challenging V3 english tasks (Qwen2.5). These consistent trends across multiple VLM backbones confirm that principled capacity regulation is essential for complex retrieval scenarios.

\subsection{Experimental Analysis}
\label{subsec:Analysis}
We analyze our framework's core mechanisms using Qwen2 on ViDoRe V1. An ablation study validates our three design objectives by comparing heuristic baselines against optimal transport. Qualitative visualizations then show how our approach preserves geometric density and prevents representation collapse. Additional results are in the supplement.

\begin{table*}[!t]
    \centering
    \tiny
    \caption{\textbf{Ablation study on ViDoRe V1 academic tasks.} nDCG@5/Recall@1 ($\uparrow$) and Time (ms/q). Best compression results per MF are bolded.}
    \vspace{-0.2cm}
    {\setlength{\tabcolsep}{3pt}
    \begin{tabular}{L{0.235\linewidth}|C{0.034\linewidth}|*{5}{C{0.0975\linewidth}|}C{0.0975\linewidth}}
        \toprule
        \textbf{Method} & 
        \textbf{MF} &
        \textbf{ArxivQ} & 
        \textbf{DocQ} & 
        \textbf{InfoQ} & 
        \textbf{TabF} & 
        \textbf{TATQ} & 
        \textbf{Avg.}\\
        \midrule
        (1) HAC: Dist & \multirow{5}{*}{25} & 82.8/76.4 & 47.4/38.4 & 72.9/63.0 & 86.5/78.2 & 52.2/38.7 & 69.3/60.3 \\
        (2) HAC: Dist+Attn & & 82.9/75.4 & 48.7/39.6 & 80.5/72.6 & 87.2/79.6 & 58.3/44.6 & 71.5/62.4 \\
        (3) HAC: Dist+Attn+Cap. &  & 83.4/75.8 & 48.3/38.6 & 81.2/73.4 & \textbf{88.5}/\textbf{82.1} & 61.3/47.4 & 72.5/63.5 \\
        (4) Balanced OT &  & \textbf{84.2}/\textbf{77.2} & 50.4/\textbf{41.4} & 84.9/77.6 & 88.1/81.4 & 67.8/54.7 & 75.1/66.5 \\
        (5) Unbalanced OT (Ours) & & 83.5/76.6 & \textbf{50.8}/\textbf{41.4} & \textbf{85.8}/\textbf{79.2} & 87.7/81.4 & \textbf{69.0}/\textbf{55.5} & \textbf{75.4}/\textbf{66.8} \\
        \midrule        
        (1) HAC: Dist & \multirow{5}{*}{50} & 75.9/67.4 & 37.8/30.2 & 65.8/55.8 & 81.3/72.5 & 35.7/25.2 & 61.6/52.3 \\
        (2) HAC: Dist+Attn & & 78.3/71.4 & 38.5/29.2 & 72.3/62.2 & 79.6/68.6 & 44.5/32.2 & 62.6/52.7 \\
        (3) HAC: Dist+Attn+Cap. & & 79.9/73.0 & 41.7/33.0 & 76.4/67.8 & 82.8/73.9 & 46.4/32.9 & 65.4/56.1 \\
        (4) Balanced OT &  & 80.7/73.0 & 44.9/35.6 & 80.9/\textbf{73.8} & \textbf{87.2}/\textbf{80.7} & \textbf{61.2}/\textbf{47.9} & 71.0/62.1 \\
        (5) Unbalanced OT (Ours) & & \textbf{82.2}/\textbf{75.4} & \textbf{45.2}/\textbf{36.4} & \textbf{81.2}/73.6 & 86.8/79.6 & 60.1/47.1 & \textbf{71.3}/\textbf{62.4} \\
        \bottomrule
    \end{tabular}}
    \label{tab:ablation_formulation}
\end{table*}

\subsubsection{Ablation Study.}
We perform this ablation study under MF $\in \{25, 50\}$, where performance disparities are most pronounced. As noted in~\cref{subsec:token_compression}, effective compression requires balancing three objectives: minimizing token redundancy, preserving token importance, and regulating coverage capacity. We incrementally integrate them into a Hierarchical Agglomerative Clustering (HAC) baseline. We heuristically model token redundancy as centroid Euclidean distance, token importance as average attention, and coverage capacity as the total attention sum per cluster. The supplementary material details these criteria.

\Cref{tab:ablation_formulation} shows that relying solely on feature redundancy (HAC: Dist) yields the lowest performance (57.9 nDCG@5 at MF=50). Incrementally integrating attention-based importance (HAC: Dist+Attn) and bounding coverage capacity (HAC: Dist+Attn+Cap.) prevents excessive merging, progressively boosting scores to 62.6 and 65.4. This confirms all three objectives are crucial for maintaining the representation diversity required by MaxSim. While these heuristics validate our design choices, their greedy nature remains sub-optimal. Unifying these objectives into an Optimal Transport (OT) framework theoretically bounds MaxSim degradation, improving performance. Our Unbalanced OT formulation relaxes target marginals; allowing representative tokens to adapt their capacities to local semantic density minimizes transport cost further, achieving the highest overall performance.

\begin{figure*}[t]
    \centering
    \begin{subfigure}[t]{0.44\linewidth}
        \centering
        \includegraphics[width=\linewidth]{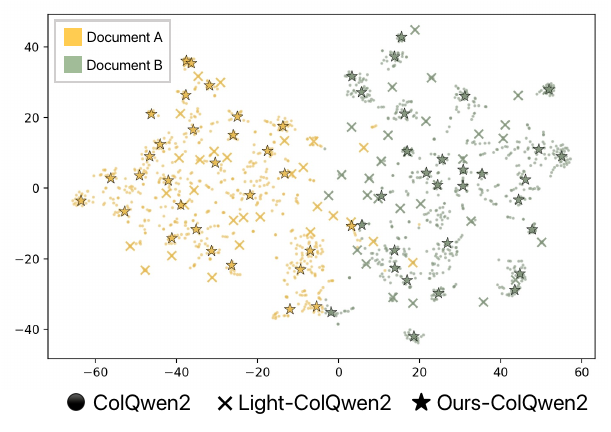}
        \caption{t-SNE.}
        \label{fig:tsne}
    \end{subfigure}
    \hfill
    \begin{subfigure}[t]{0.48\linewidth}
        \centering
        \includegraphics[width=\linewidth]{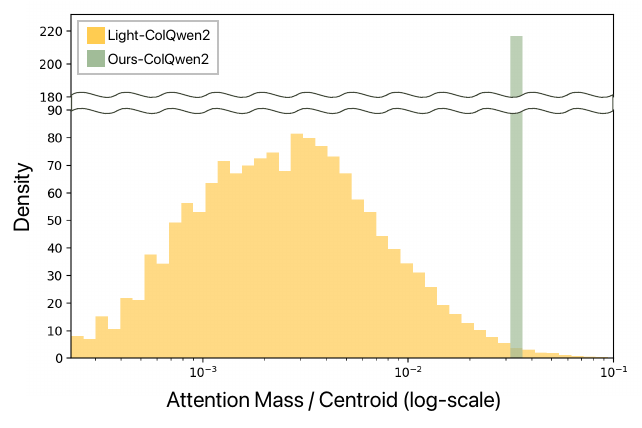}
        \caption{Attention histogram.}
        \label{fig:histogram}
    \end{subfigure}
    \caption{\textbf{Qualitative analysis (MF=25).} (a) t-SNE: embedding regions of original tokens ($\bullet$), Light-ColQwen2 ($\times$), and Ours ($\bigstar$). (b) Attention histogram: distribution of mass per representative token; green bar indicates the uniform share ($\approx 0.033$).}
    \label{fig:qualitative_analysis}
\end{figure*}

\subsubsection{Geometric Preservation in Embedding Space.}
\Cref{fig:tsne} visualizes token embeddings via t-SNE\cite{van2008visualizing} at MF=25. While original ColQwen2 tokens ($\bullet$) form dense semantic regions, Light-ColQwen2 representatives ($\times$) often deviate into low-density areas because unconstrained clustering skews large cluster centroids and overfits small ones. In contrast, our UOT anchors ($\bigstar$) remain within the original high-density regions. This aligns with ~\cref{thm:worst_case}'s geometric covering guarantee: penalizing the transport cost restricts representatives near their source tokens, thereby minimizing semantic distortion and preserving accurate MaxSim retrieval.

\subsubsection{Attention Mass Distribution.}
\Cref{fig:histogram} shows the density histogram of attention mass per representative token (log-scale). Light-ColQwen2 yields a highly skewed distribution where a few centroids dominate, indicating representation collapse from unconstrained merging. Conversely, our capacity regulation distributes mass more uniformly, successfully concentrating it near the expected uniform share ($\approx 0.033$ for $K=30$ tokens at MF=25). This ensures all compressed tokens contribute effectively to the semantic coverage, minimizing redundancy and dead capacity.

\section{Conclusion}
\label{sec:conclusion}
This work addressed the scalability bottleneck in VLM-based Visual Document Retrieval caused by the token explosion in late-interaction models. We demonstrated that effective token compression requires the principled integration of semantic redundancy, token importance, and coverage capacity. By reformulating this task as an Unbalanced Optimal Transport (UOT) problem, we proposed a unifying framework that optimizes these objectives within a single formulation. Supported by theoretical guarantees that bound MaxSim approximation errors, our approach establishes a superior efficiency–effectiveness trade-off on the ViDoRe V1 and V3 benchmarks, consistently outperforming state-of-the-art baselines across diverse VLM architectures.

\section*{Acknowledgements}
This work was supported in part by the National Research Foundation of Korea (NRF) grant funded by the Korean Government (Grant No. RS-2024-00352566), the Institute of Information \& Communications Technology Planning \& Evaluation (IITP) grant funded by the Ministry of Science and ICT (MSIT) (Grant No. RS-2026-25510490), and Seoul National University of Science and Technology.

%
%
\bibliographystyle{splncs04}
\bibliography{main}
\end{document}


\title{Unbalanced Optimal Transport for Efficient Visual Document Retrieval\\--Supplementary Materials--} 

\titlerunning{Unbalanced Optimal Transport for Efficient Visual Document Retrieval}

\author{
Hoyeon Shin\inst{1}$^{\star}$\orcidlink{0009-0007-8016-5022} 
\and
Jeongyeon Kim\inst{1,2}\thanks{Equal contribution.}\orcidlink{0009-0004-1520-6201} 
\and
Yeong Jun Koh\inst{3}\orcidlink{0000-0003-1805-2960}
\and 
Yeoneung Kim\inst{1}\orcidlink{0000-0001-6293-3939}
\and 
Hanul Kim\inst{1}\thanks{Corresponding author.}\orcidlink{0000-0001-7450-6600}}
\authorrunning{H.~Shin et al.}

\institute{Seoul National University of Science and Technology, Seoul, Korea
\email{\{hy\_shin,jy\_kim, yeoneung, hukim\}@seoultech.ac.kr}
\and
SIZL, Seongnam, Korea
\and
Chungnam National University, Daejeon, Korea
\email{\{yjkoh\}@cnu.ac.kr}
}



\maketitle

\setcounter{section}{0}
\setcounter{figure}{0}
\setcounter{table}{0}
\setcounter{equation}{0}

\renewcommand{\thesection}{\Alph{section}}

\renewcommand{\thefigure}{S\arabic{figure}}
\renewcommand{\thetable}{S\arabic{table}}
\renewcommand{\theequation}{S\arabic{equation}}
\renewcommand{\thealgorithm}{S\arabic{algorithm}}

This supplementary material provides further details, mathematical proofs, and extended empirical evaluations to support the main manuscript. The document is organized as follows:

\begin{itemize}
    \item \cref{sec:supp_empirical_validation} presents additional experimental results. This includes empirical validations of our theoretical bounds, an analysis of ranking preservation under compression, additional datasets evaluation, hyperparameter sensitivity studies, and a detailed comparison of indexing efficiency.
    \item \cref{sec:supp_proofs} provides the complete mathematical proofs for Theorems 1--3 introduced in the main paper, alongside auxiliary lemmas that establish the two-sided score deviation bound.
    \item \cref{sec:supp_more_theory} presents additional theoretical analysis, specifically detailing the asymptotic behavior of the marginal relaxation parameter.
\end{itemize}

\section{Additional Experimental Analysis}
\label{sec:supp_empirical_validation}

\subsubsection{Empirical vs. Theoretical MaxSim Degradation.}
\begin{figure*}[!ht]
    \centering
    \begin{subfigure}[t]{\linewidth}
        \centering
        \includegraphics[width=\linewidth]{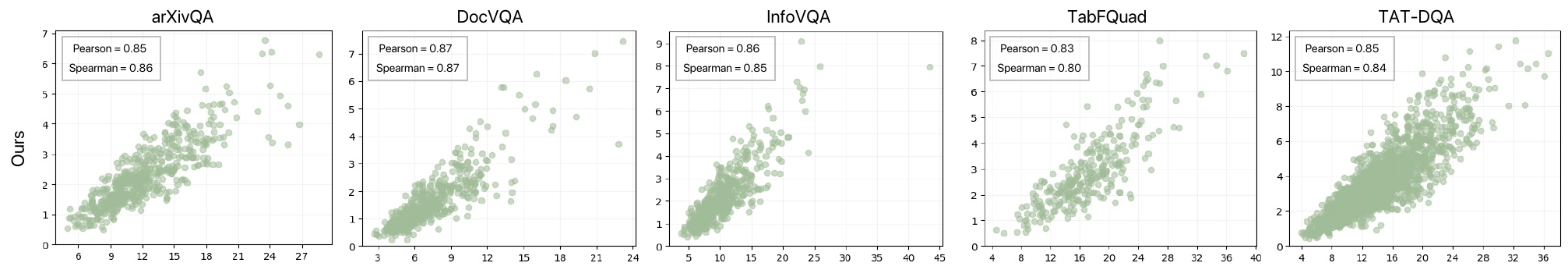}
        \caption{MF=25, Ours}
    \end{subfigure}
    \begin{subfigure}[t]{\linewidth}
        \centering
        \includegraphics[width=\linewidth]{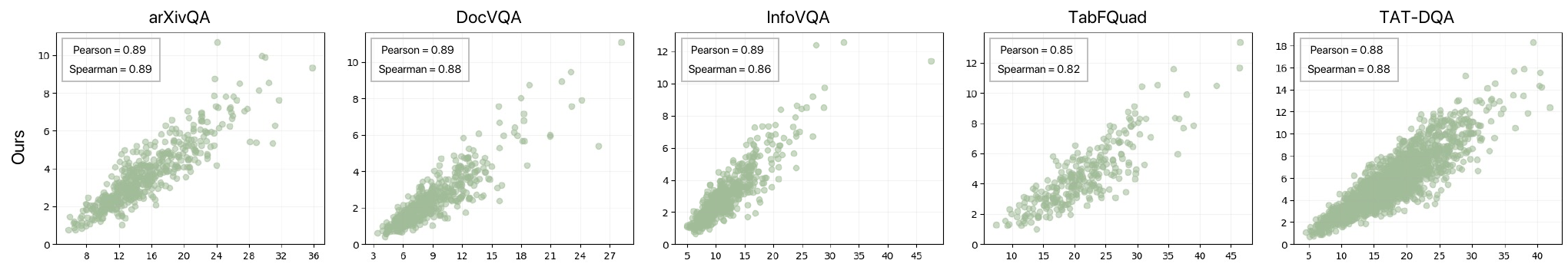}
        \caption{MF=50, Ours}
    \end{subfigure}
    \begin{subfigure}[t]{\linewidth}
        \centering
        \includegraphics[width=\linewidth]{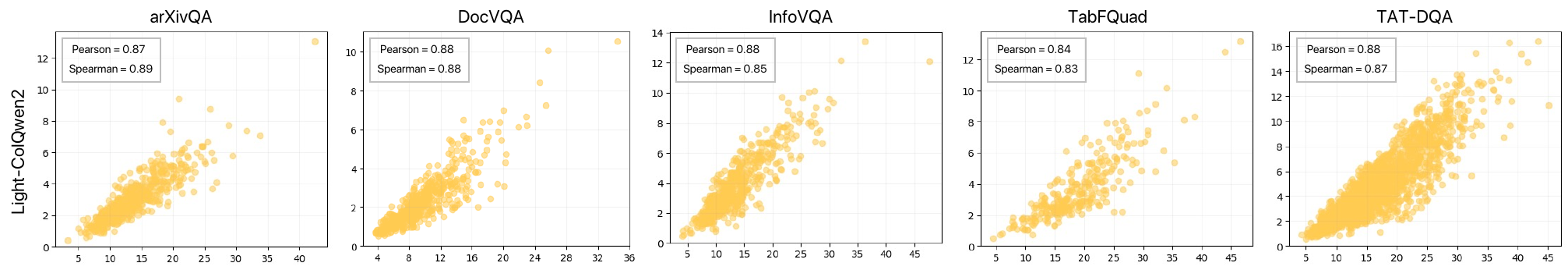}
        \caption{MF=25, Light-ColQwen2}
    \end{subfigure}
    \begin{subfigure}[t]{\linewidth}
        \centering
        \includegraphics[width=\linewidth]{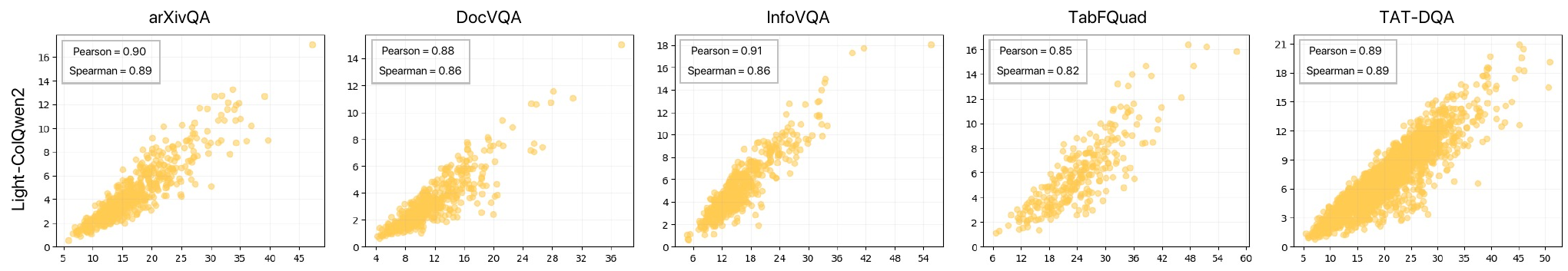}
        \caption{MF=50, Light-ColQwen2}
    \end{subfigure}
    \caption{\textbf{Empirical MaxSim Degradation vs. Theoretical Bound.} 
     Results are obtained using ColQwen2 on the ViDoRe V1 academic tasks, where each point represents a query. The y-axis shows the observed MaxSim degradation, and the x-axis corresponds to the error bound.}
     \label{fig:supp_theorem1}
\end{figure*}

To validate Theorem 1, we compute the observed MaxSim degradation, $s(\mathbf{Q},\mathbf{X})-s(\mathbf{Q},\mathbf{Y})$, for each query and compare it against the theoretical bound. \cref{fig:supp_theorem1} illustrates this relationship on the ViDoRe V1 academic tasks at compression ratios of MF=25 and 50. To demonstrate the generality of this bound, we evaluate it using two compression paradigms: our method, which learns a soft transport plan via Unbalanced Optimal Transport (UOT), and Light-ColQwen2, which restricts the transport plan to a hard binary $N \times K$ assignment matrix. Across all evaluated tasks, the predicted bound closely tracks the observed degradation, achieving high Pearson (0.87) and Spearman (0.83) correlations. This consistent agreement across soft and hard plans demonstrates that transport cost reliably predicts score deviation. Thus, the UOT objective effectively controls this bound, preserving retrieval stability.

\subsubsection{Empirical Worst-Case MaxSim Degradation vs. Theoretical Bound.}
\begin{figure*}[!ht]
    \centering
    \begin{subfigure}[t]{\linewidth}
        \centering
        \includegraphics[width=\linewidth]{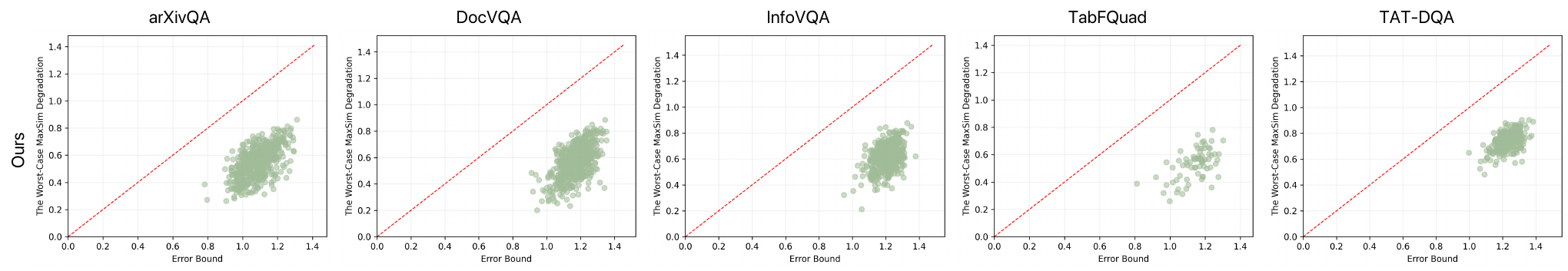}
        \caption{MF=25, Ours}
    \end{subfigure}
    \begin{subfigure}[t]{\linewidth}
        \centering
        \includegraphics[width=\linewidth]{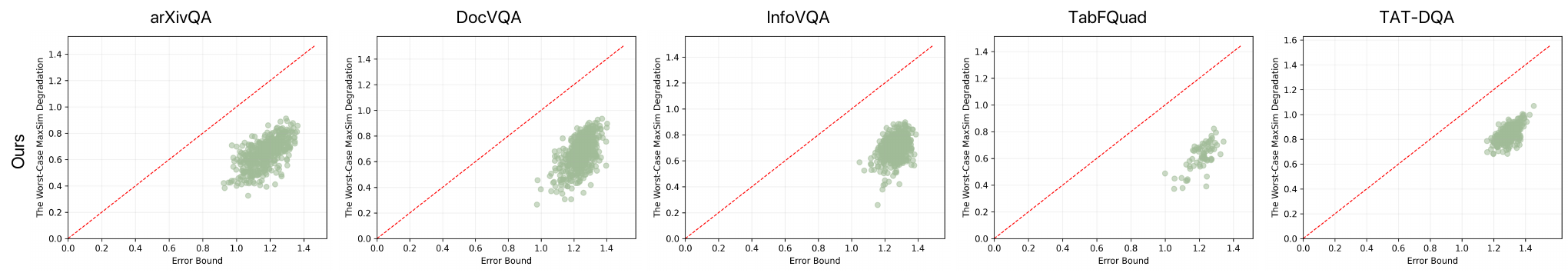}
        \caption{MF=50, Ours}
    \end{subfigure}
    \begin{subfigure}[t]{\linewidth}
        \centering
        \includegraphics[width=\linewidth]{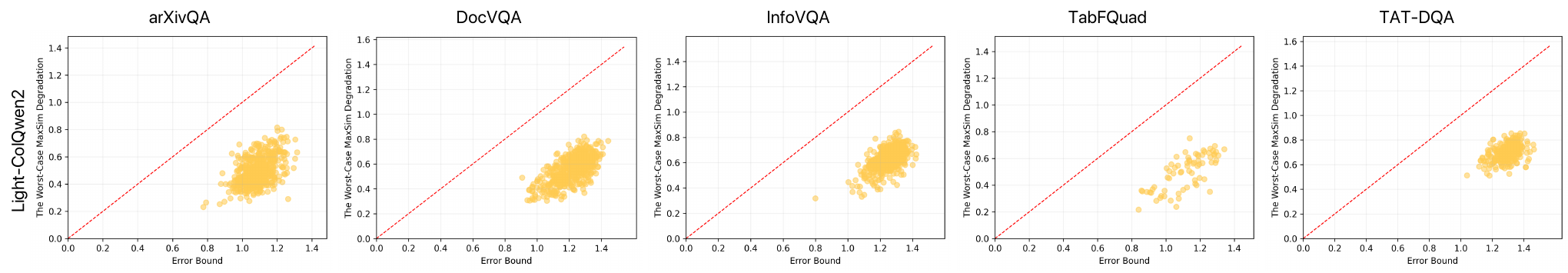}
        \caption{MF=25, Light-ColQwen2}
    \end{subfigure}
    \begin{subfigure}[t]{\linewidth}
        \centering
        \includegraphics[width=\linewidth]{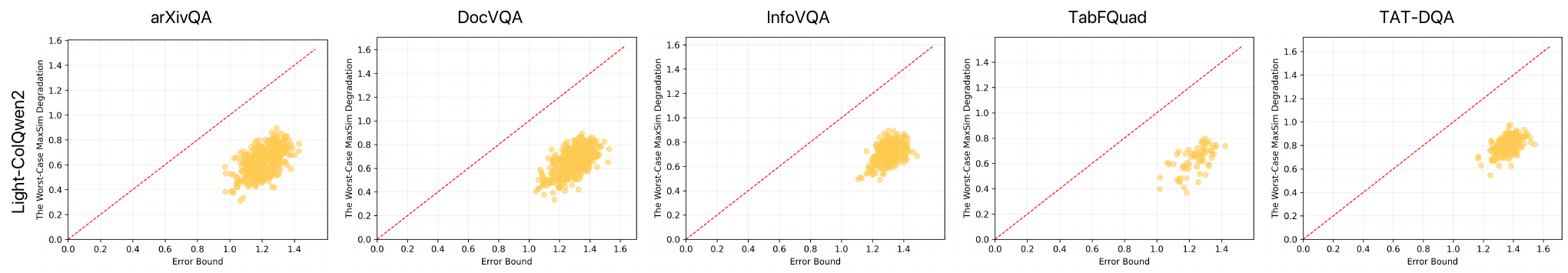}
        \caption{MF=50, Light-ColQwen2}
    \end{subfigure}
    \caption{\textbf{Empirical Worst-Case MaxSim Degradation vs. Theoretical Bound.} 
     Results are obtained using ColQwen2 on the ViDoRe V1 academic tasks, where each point represents a document. The y-axis shows the worst-case MaxSim degradation, and the x-axis corresponds to the error bound.}
     \label{fig:supp_theorem2}
\end{figure*}

To validate Theorem 2, we evaluate the worst-case MaxSim degradation, $s(\mathbf{Q},\mathbf{X})-s(\mathbf{Q},\mathbf{Y})$, for each document by computing the maximum score drop against all dataset query tokens. \cref{fig:supp_theorem2} visualizes this empirical maximum and our theoretical upper bound for both our method and Light-ColQwen2 on the ViDoRe V1 academic tasks at compression ratios of MF=25 and 50. In all cases, the theoretical limit strictly upper-bounds the observed worst-case degradation with zero violations. Although conservative due to worst-case geometric assumptions, the bound closely tracks the scale and trend of the empirical errors, confirming its reliability as a strict structural safety net for our compression framework.

\subsubsection{Empirical Ranking Preservation.}
\begin{figure*}[!ht]
    \centering
    \begin{subfigure}[t]{\linewidth}
        \centering
        \includegraphics[width=\linewidth]{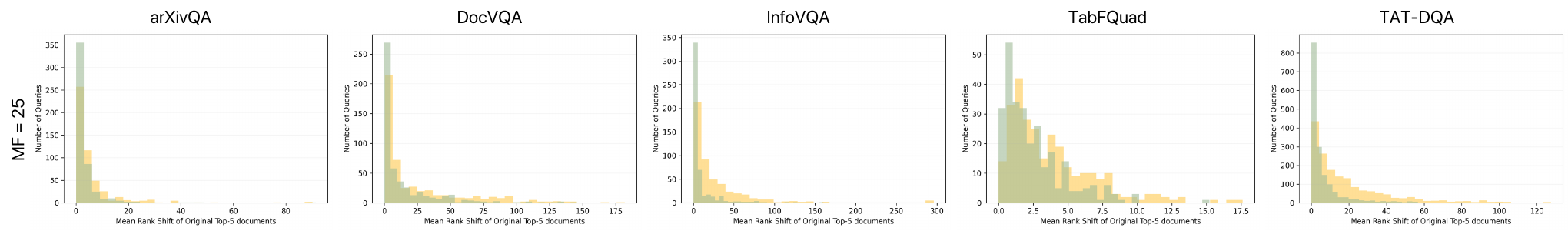}
        \caption{Top-5, MF=25}
    \end{subfigure}
    \begin{subfigure}[t]{\linewidth}
        \centering
        \includegraphics[width=\linewidth]{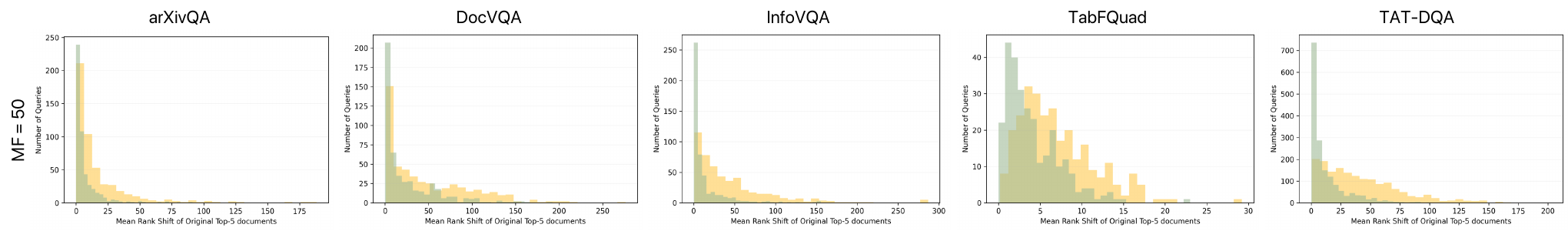}
        \caption{Top-5, MF=50}
    \end{subfigure}
    \begin{subfigure}[t]{\linewidth}
        \centering
        \includegraphics[width=\linewidth]{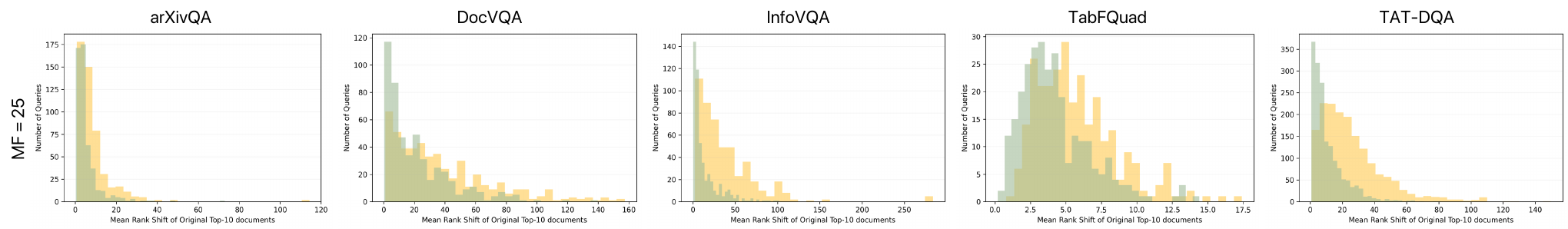}
        \caption{Top-10, MF=25}
    \end{subfigure}
    \begin{subfigure}[t]{\linewidth}
        \centering
        \includegraphics[width=\linewidth]{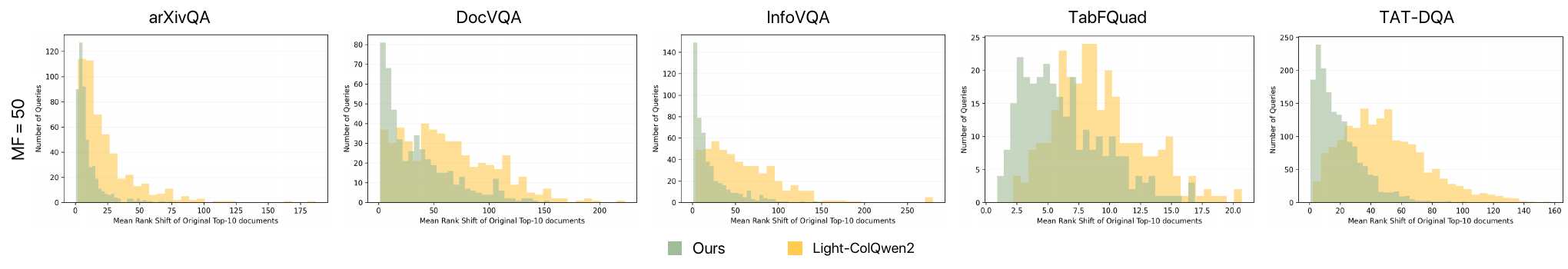}
        \caption{Top-10, MF=50}
    \end{subfigure}
    \caption{\textbf{Distribution of Ranking Preservation under Compression.} 
    Results are obtained using ColQwen2 on the ViDoRe V1 academic tasks at compression ratios of MF=25 and 50. We compare the ranking stability of our method against Light-ColQwen2. The x-axis represents the query-wise mean absolute rank shift of the documents that originally belonged to the top-$k$ set ($k \in \{5, 10\}$), while the y-axis shows the number of queries.}
    \label{fig:supp_theorem3}
\end{figure*}

To empirically validate the ranking stability, we first identify the top-$k$ documents ($k \in \{5, 10\}$) retrieved by the uncompressed baseline. For each query, we then track these exact documents within the compressed model's ranked list and compute their mean absolute rank shift. \cref{fig:supp_theorem3} compares the distribution of these rank shifts between our method and Light-ColQwen2 on the ViDoRe V1 academic tasks at compression ratios of MF=25 and 50. Across all datasets and compression levels, Light-ColQwen2 exhibits substantially larger rank perturbations, with long right-tailed distributions indicating that many originally high-ranked documents are displaced far down the ranking. In contrast, our method yields distributions tightly concentrated near zero, successfully preserving the original top-$k$ positions. Crucially, this performance gap widens under extreme compression (MF=50): while Light-ColQwen2 suffers from exacerbated rank shifts, our approach maintains robust ranking stability.

\subsubsection{Hyperparameter Sensitivity to $\gamma$.}
\begin{figure*}[!ht]
    \centering
    \begin{subfigure}[t]{0.49\linewidth}
        \centering
        \includegraphics[width=\linewidth,height=4.5cm]{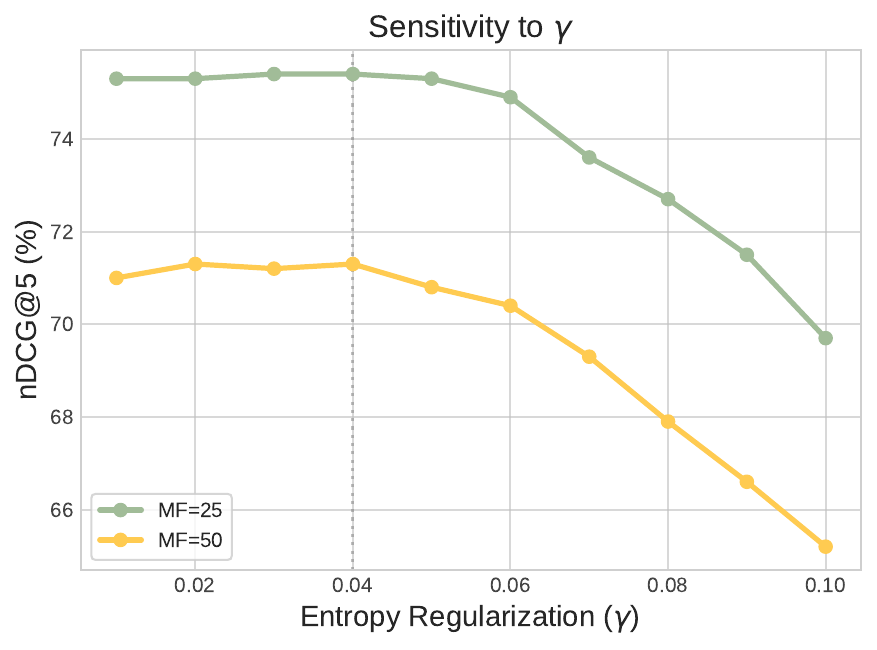}
        \caption{}
    \end{subfigure}
    \hfill
    \begin{subfigure}[t]{0.49\linewidth}
        \centering
        \includegraphics[width=\linewidth,height=4.5cm]{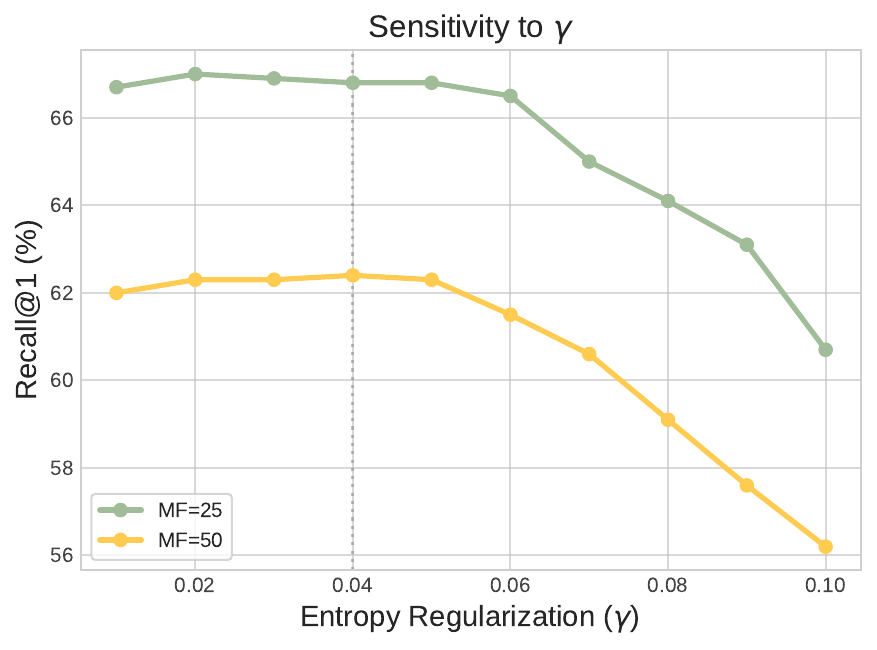}
        \caption{}
    \end{subfigure}
    \caption{\textbf{Hyperparameter Sensitivity to $\gamma$.} 
     Results are obtained using ColQwen2 on the ViDoRe V1 academic tasks.}
     \label{fig:supp_hyperparam_gamma}
\end{figure*}

The parameter $\gamma$ controls the smoothness of the transport plan. Retrieval accuracy remains stable for $\gamma \in [0.02, 0.05]$ (\cref{fig:supp_hyperparam_gamma}), demonstrating robustness within this range. Increasing $\gamma \ge 0.06$ degrades performance because excessive entropy forces the transport plan toward a uniform distribution, obscuring precise token-to-token alignments. Conversely, low values ($\gamma \le 0.01$) provide insufficient regularization, hindering optimization and yielding suboptimal results. We set $\gamma=0.04$ by default to ensure adequate smoothing without compromising alignment precision.

\subsubsection{Additional Datasets Evaluation.}
In addition to the main experiments, we evaluate our method on ViDoRe V2 and the cross-lingual split of ViDoRe V3. ViDoRe V2 is included as an additional benchmark to verify the generality of our method beyond the main evaluation suite. We also report cross-lingual ViDoRe V3 results using multilingual queries, including English, to examine whether the proposed compression framework remains effective under multilingual retrieval scenarios. We restrict experiments under MF $\in {5, 10, 25,50}$ using only ColQwen2 and ColQwen2.5.

As shown in \cref{tab:sub_v3_cross,tab:sub_v2}, our method preserves retrieval quality well across both benchmarks, particularly under aggressive compression. On ViDoRe V3 cross-lingual tasks, our method consistently outperforms Light-ColQwen at MF=50, improving average nDCG@5 by +11.8 with Qwen2 and +19.0 with Qwen2.5. Similar trends are observed on ViDoRe V2, where our method achieves substantial gains over Light-ColQwen at MF=50, with +13.9 and +23.7 average nDCG@5 improvements for Qwen2 and Qwen2.5, respectively. These results further demonstrate that the proposed capacity-regulated compression strategy generalizes well beyond the main benchmarks and remains effective in multilingual and diverse document retrieval scenarios.

\begin{table*}[!ht]
    \centering
    \tiny 
    \caption{\textbf{Results on ViDoRe V2.} nDCG@5/Recall@1 ($\uparrow$) and Time (ms/q). Best compression results per MF are bolded.}
    \vspace{-0.2cm}
    {\setlength{\tabcolsep}{3.2pt}
    \begin{tabular}{L{0.18\linewidth}|C{0.035\linewidth}|C{0.065\linewidth}|*{4}{C{0.11\linewidth}|}C{0.11\linewidth}}
        \toprule
        \textbf{Method} & 
        \textbf{MF} & 
        \textbf{Time} & 
        \textbf{Bio.} & 
        \textbf{Econ.} & 
        \textbf{ESG} & 
        \textbf{ESG H.} & 
        \textbf{Avg.}\\
        \midrule

        \rowcolor{lightgray!35}
        \multicolumn{8}{l}{\textit{Base model: Qwen2-VL-2B}}\\
        \midrule
        ColQwen2 & - &1.80 & 56.6/53.9   & 54.3/57.8  &  54.3/50.4&  57.2/57.7& 55.6/55.0 \\
        \midrule

        ColQwen2+Prune & \multirow{3}{*}{5} & \multirow{3}{*}{0.28} 
        & \textbf{55.1}/\textbf{54.2} 
        & 48.1/55.6 
        & 50.0/45.6 
        & \textbf{53.2}/\textbf{50.0} 
        & 51.6/\textbf{51.4} \\
        
        Light-ColQwen2 &  & 
        & 54.0/51.1 
        & 47.4/54.7 
        & 51.7/49.1 
        & 49.8/48.1 
        & 50.7/50.8  \\
        Ours &  &
        & 54.6/51.2 
        & \textbf{51.8}/\textbf{56.9} 
        & \textbf{53.7}/\textbf{50.0} 
        & 47.8/40.4 
        & \textbf{52.0}/49.6 \\
        \midrule

        ColQwen2+Prune & \multirow{3}{*}{10} & \multirow{3}{*}{0.19} 
        & 50.5/49.4  
        & 47.3/\textbf{55.2} 
        & 45.1/41.7 
        & 44.8/36.5 
        & 46.9/45.7 \\

        Light-ColQwen2 &  & 
        & 52.1/\textbf{49.8}  
        & 40.4/48.7 
        & 46.5/42.1 
        & 40.8/30.8 
        & 45.0/42.9 \\
        Ours &  &
        & \textbf{53.4}/\textbf{49.8} 
        & \textbf{50.4}/\textbf{55.2} 
        & \textbf{52.3}/\textbf{47.4} 
        & \textbf{48.2}/\textbf{42.3} 
        & \textbf{51.1}/\textbf{48.7} \\
        \midrule

        ColQwen2+Prune & \multirow{3}{*}{25} & \multirow{3}{*}{0.14} 
        & 45.0/42.0  
        & 44.4/51.7 
        & 31.7/30.3 
        & 40.8/\textbf{34.6} 
        & 40.5/39.7 \\

        Light-ColQwen2 & & 
        & 46.9/43.8	
        & 35.6/43.1 
        & 36.1/34.6	
        & 29.9/21.2	
        & 37.1/35.7\\
        Ours &  &
        & \textbf{51.0}/\textbf{47.2} 
        & \textbf{49.7}/\textbf{53.0} 
        & \textbf{43.4}/\textbf{39.9} 
        & \textbf{44.6}/\textbf{34.6} 
        & \textbf{47.2}/\textbf{43.7}\\
        \midrule

        ColQwen2+Prune & \multirow{3}{*}{50} & \multirow{3}{*}{0.13} 
        & 37.2/34.2   
        & 38.6/40.1 
        & 26.0/26.3 
        & 26.9/23.1
        & 32.2/30.9 \\

        Light-ColQwen2 &  &
        & 42.1/39.8
        & 32.4/38.4
        & 28.9/22.4	
        & 17.9/7.7
        & 30.3/27.1\\
        Ours &  &
        & \textbf{47.1}/\textbf{43.3} 
        & \textbf{46.4}/\textbf{51.3} 
        & \textbf{40.0}/\textbf{31.1} 
        & \textbf{43.3}/\textbf{32.7} 
        & \textbf{44.2}/\textbf{39.6}\\
        \midrule

        \rowcolor{lightgray!35}
        \multicolumn{8}{l}{\textit{Base model: Qwen2.5-VL-3B}}\\
        \midrule
        ColQwen2.5 & - & 1.04
        & 57.5/53.1  & 52.9/55.6  & 50.4/43.9 & 64.1/65.4 & 56.2/54.5 \\
        \midrule

        ColQwen2.5+Prune & \multirow{3}{*}{5} & \multirow{3}{*}{0.27} 
        & 54.6/48.1 
        & \textbf{51.2}/\textbf{52.6} 
        & 47.9/\textbf{44.7} 
        & 44.2/32.7 
        & 49.5/44.5 \\
        
        Light-ColQwen2.5 &  &
        & 54.5/49.5 
        & 38.7/45.7 
        & 38.1/32.0 
        & 44.2/38.5 
        & 43.9/41.4 \\
        Ours &  &
        & \textbf{56.6}/\textbf{51.4} 
        & 49.7/50.4 
        & \textbf{48.5}/42.1 
        & \textbf{57.7}/\textbf{50.0} 
        & \textbf{53.1}/\textbf{48.5} \\
        \midrule

        ColQwen2.5+Prune & \multirow{3}{*}{10} & \multirow{3}{*}{0.18} 
        & 51.8/46.1 
        & \textbf{50.2}/\textbf{55.6} 
        & \textbf{45.6}/38.6 
        & 42.8/34.6 
        & 47.6/43.7 \\

        Light-ColQwen2.5 & & 
        & 50.9/46.4 
        & 27.1/30.6 
        & 32.9/25.4 
        & 32.4/26.9 
        & 35.8/32.3 \\
        Ours &  &
        & \textbf{54.6}/\textbf{51.2} 
        & 49.4/48.3 
        & 43.8/\textbf{39.0} 
        & \textbf{55.4}/\textbf{53.8} 
        & \textbf{50.8}/\textbf{48.1} \\
        \midrule

        ColQwen2.5+Prune & \multirow{3}{*}{25} & \multirow{3}{*}{0.14} 
        & 44.9/40.9 
        & 43.0/48.3 
        & 31.7/32.5 
        & 34.2/32.7 
        & 38.5/38.6  \\

        Light-ColQwen2.5 &  &
        & 39.1/34.2 
        & 22.2/26.3 
        & 18.4/15.4 
        & 20.4/17.3 
        & 25.0/23.3 \\
        Ours &  &
        & \textbf{52.5}/\textbf{49.2} 
        & \textbf{48.4}/\textbf{49.6} 
        & \textbf{42.9}/\textbf{38.2} 
        & \textbf{43.7}/\textbf{44.2} 
        & \textbf{46.9}/\textbf{45.3} \\
        \midrule

        ColQwen2.5+Prune & \multirow{3}{*}{50} & \multirow{3}{*}{0.12}  
        & 39.8/37.2 
        & 34.6/44.4 
        & 23.5/21.5 
        & 32.3/32.7 
        & 32.6/34.0\\

        Light-ColQwen2.5 &  & 
        & 27.8/22.2 
        & 20.9/22.0 
        & 12.6/8.8  
        & 18.7/13.5 
        & 20.0/16.6 \\
        Ours &  &
        & \textbf{48.6}/\textbf{46.1} 
        & \textbf{47.7}/\textbf{51.3} 
        & \textbf{36.6}/\textbf{29.8} 
        & \textbf{41.7}/\textbf{38.5} 
        & \textbf{43.7}/\textbf{41.4} \\
        \bottomrule
    \end{tabular}}
    \label{tab:sub_v2}
\end{table*}

\begin{table*}[!t]
    \centering
    \tiny 
    \caption{\textbf{Results on ViDoRe V3 cross-lingual tasks.} nDCG@5/Recall@1 ($\uparrow$) and Time (ms/q). Best compression results per MF are bolded.}
    \vspace{-0.2cm}
    {\setlength{\tabcolsep}{3pt}
    \begin{tabular}{L{0.169\linewidth}|C{0.035\linewidth}|C{0.060\linewidth}|*{5}{C{0.0975\linewidth}|}C{0.0975\linewidth}}
        \toprule
        \textbf{Method} & 
        \textbf{MF} & 
        \textbf{Time} & 
        \textbf{C.S.} & 
        \textbf{Fin.} & 
        \textbf{H.R.} & 
        \textbf{Ind.} & 
        \textbf{Phar.} & 
        \textbf{Avg.}\\
        \midrule

        \rowcolor{lightgray!35}
        \multicolumn{9}{l}{\textit{Base model: Qwen2-VL-2B}}\\
        \midrule
        ColQwen2 & - &  3.08& 62.2/67.8 & 45.5/47.4 & 46.7/47.5 & 36.5/36.0 & 56.8/57.9 & 49.5/51.3 \\
        \midrule
        ColQwen2+Prune & \multirow{3}{*}{5} & \multirow{3}{*}{0.76} 
        & \textbf{63.0}/\textbf{69.6} 
        & \textbf{43.0}/44.3 
        & 44.4/\textbf{45.4} 
        & 35.8/35.5 
        & 51.9/52.5 
        & 47.6/\textbf{49.5} \\
        
        Light-ColQwen2 &  & 
        &62.2/68.2&39.9/40.8&42.1/41.6&36.5/35.7&52.2/51.4&46.6/47.5\\
        Ours &  &
        & 62.4/67.8&42.8/\textbf{44.6}&\textbf{45.1}/44.9&\textbf{37.1}/\textbf{37.0}&\textbf{53.2}/\textbf{52.6}&\textbf{48.1}/49.4\\
        \midrule

        ColQwen2+Prune & \multirow{3}{*}{10} & \multirow{3}{*}{0.47} 
        & 57.4/61.8 
        & 36.8/38.2 
        & 41.3/42.5 
        & 29.1/28.9 
        & 47.6/46.5 
        & 42.4/43.6 \\

        Light-ColQwen2 &  & 
        &59.9/66.7&36.6/37.3&38.3/38.5&33.3/32.3&50.9/50.0&43.8/45.0\\
        Ours &  &
        &\textbf{61.6}/\textbf{67.9}&\textbf{40.2}/\textbf{40.2}&\textbf{43.2}/\textbf{42.7}&\textbf{35.4}/\textbf{34.0}&\textbf{52.1}/\textbf{51.4}&\textbf{46.5}/\textbf{47.3}\\
        \midrule

        ColQwen2+Prune & \multirow{3}{*}{25} & \multirow{3}{*}{0.32} 
        & 52.3/56.4 
        & 27.5/28.8 
        & 33.5/34.7 
        & 23.9/21.8 
        & 38.3/38.6 
        & 35.1/36.1 \\

        Light-ColQwen2 & & 
        &53.4/59.4&27.1/30.0&29.5/32.0&25.9/24.6&45.2/44.7&36.2/38.1\\
        Ours &  &
        &\textbf{58.7}/\textbf{64.9}&\textbf{35.3}/\textbf{37.9}&\textbf{39.6}/\textbf{39.0}&\textbf{31.4}/\textbf{29.5}&\textbf{49.6}/\textbf{48.5}&\textbf{42.9}/\textbf{43.9}\\
        \midrule

        ColQwen2+Prune & \multirow{3}{*}{50} & \multirow{3}{*}{0.23} 
        & 45.7/50.5 
        & 20.2/21.3 
        & 27.2/28.7 
        & 19.8/18.4 
        & 31.3/30.3 
        & 28.8/29.8 \\

        Light-ColQwen2 &  &
        &44.3/50.3&15.6/15.5&21.2/22.2&19.2/19.1&34.3/34.8&26.9/28.4\\
        Ours &  &
        &\textbf{55.5}/\textbf{61.1}&\textbf{29.5}/\textbf{30.7}&\textbf{36.9}/\textbf{38.2}&\textbf{27.7}/\textbf{26.2}&\textbf{43.8}/\textbf{43.9}&\textbf{38.7}/\textbf{40.0}\\
        \midrule

        \rowcolor{lightgray!35}
        \multicolumn{9}{l}{\textit{Base model: Qwen2.5-VL-3B}}\\
        \midrule
        ColQwen2.5 & -  &3.05
        & 65.0/71.8 & 42.9/44.6 & 45.0/44.7 & 35.5/36.8& 55.1/54.8 & 48.7/50.5 \\
        \midrule
        ColQwen2.5+Prune & \multirow{3}{*}{5} & \multirow{3}{*}{0.74}  & 59.3/66.0 
        & 38.6/41.4 
        & \textbf{43.8}/\textbf{44.4} 
        & 29.6/28.7 
        & 52.2/52.6 
        & 44.7/46.6 \\

        Light-ColQwen2.5 &  &
        & 58.7/63.7 & 33.7/33.6 & 37.5/37.3 & 30.6/31.2 & 53.0/52.9 & 42.1/43.7 \\
        Ours &  &
        & \textbf{64.3}/\textbf{70.4} 
        & \textbf{42.0}/\textbf{45.0} 
        & 43.6/43.9 
        & \textbf{34.2}/\textbf{35.2} 
        & \textbf{54.1}/\textbf{53.9} 
        & \textbf{47.6}/\textbf{49.7} \\
        \midrule

        ColQwen2.5+Prune & \multirow{3}{*}{10} & \multirow{3}{*}{0.46} 
        & 55.4/61.4 & 32.7/35.8 & 37.4/38.3 & 25.4/24.3 & 48.2/47.4 & 39.8/41.4 \\

        Light-ColQwen2.5 &  & & 
        49.9/54.3 & 25.9/26.0 & 29.3/30.1 & 23.8/23.7 & 49.4/49.8 & 35.7/36.8 \\
        Ours &  &
        & \textbf{63.4}/\textbf{69.1} 
        & \textbf{39.8}/\textbf{42.0} 
        & \textbf{42.2}/\textbf{43.6} 
        & \textbf{32.8}/\textbf{32.4} 
        & \textbf{53.1}/\textbf{52.8} 
        & \textbf{46.3}/\textbf{48.0} \\
        \midrule

        ColQwen2.5+Prune & \multirow{3}{*}{25} & \multirow{3}{*}{0.31}  
        & 47.9/53.6 & 27.2/28.9 & 27.3/27.7 & 19.7/18.7 & 41.0/40.1 & 32.6/33.8 \\

        Light-ColQwen2.5 &  &
        & 36.1/41.1 & 17.9/18.5 & 16.6/17.4 & 13.5/13.9 & 39.1/39.0 & 24.6/26.0 \\
        Ours &  &
        & \textbf{61.7}/\textbf{67.2} 
        & \textbf{35.1}/\textbf{37.3} 
        & \textbf{37.8}/\textbf{39.4} 
        & \textbf{28.7}/\textbf{29.3} 
        & \textbf{49.9}/\textbf{50.0} 
        & \textbf{42.6}/\textbf{44.6} \\
        \midrule

        ColQwen2.5+Prune & \multirow{3}{*}{50} & \multirow{3}{*}{0.24}  
        & 41.2/47.7 & 20.1/20.1 & 18.8/20.3 & 12.6/12.0 & 33.0/32.6 & 25.1/26.5 \\

        Light-ColQwen2.5 &  & 
        & 28.5/31.5 & 13.1/12.5 & 13.4/12.3 & 9.3/9.7 & 29.4/28.7 & 18.8/18.9 \\
        Ours &  &
        & \textbf{56.3}/\textbf{62.0} 
        & \textbf{30.9}/\textbf{31.3} 
        & \textbf{33.2}/\textbf{34.9} 
        & \textbf{23.2}/\textbf{22.3} 
        & \textbf{45.4}/\textbf{44.9} 
        & \textbf{37.8}/\textbf{39.1} \\
        \bottomrule
    \end{tabular}}
    \label{tab:sub_v3_cross}
\end{table*}

\subsubsection{Hyperparameter Sensitivity to $\rho$.}
\begin{figure*}[!h]
    \centering
    \begin{subfigure}[t]{0.49\linewidth}
        \centering
        \includegraphics[width=\linewidth,height=4.5cm]{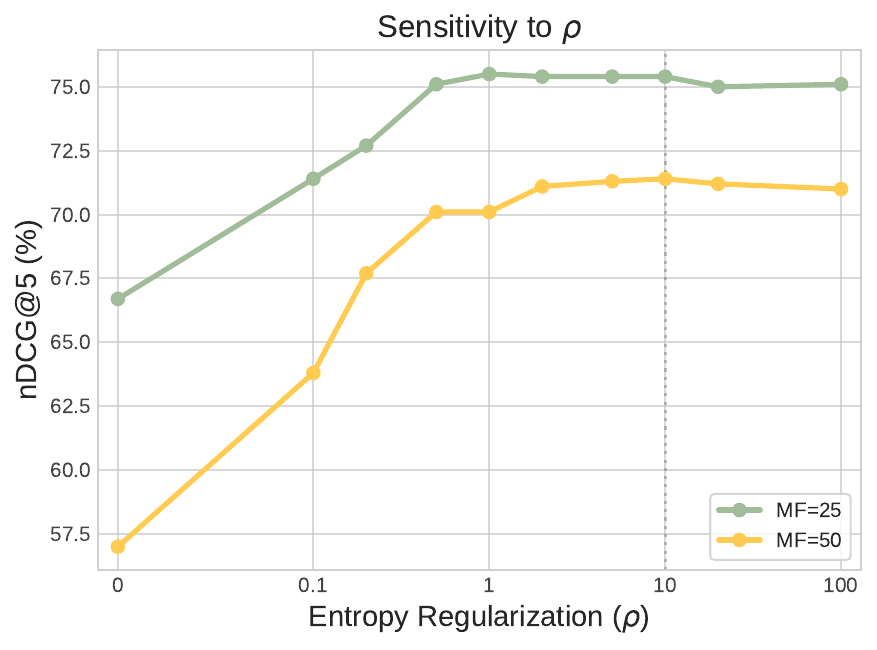}
        \caption{}
    \end{subfigure}
    \hfill
    \begin{subfigure}[t]{0.49\linewidth}
        \centering
        \includegraphics[width=\linewidth,height=4.5cm]{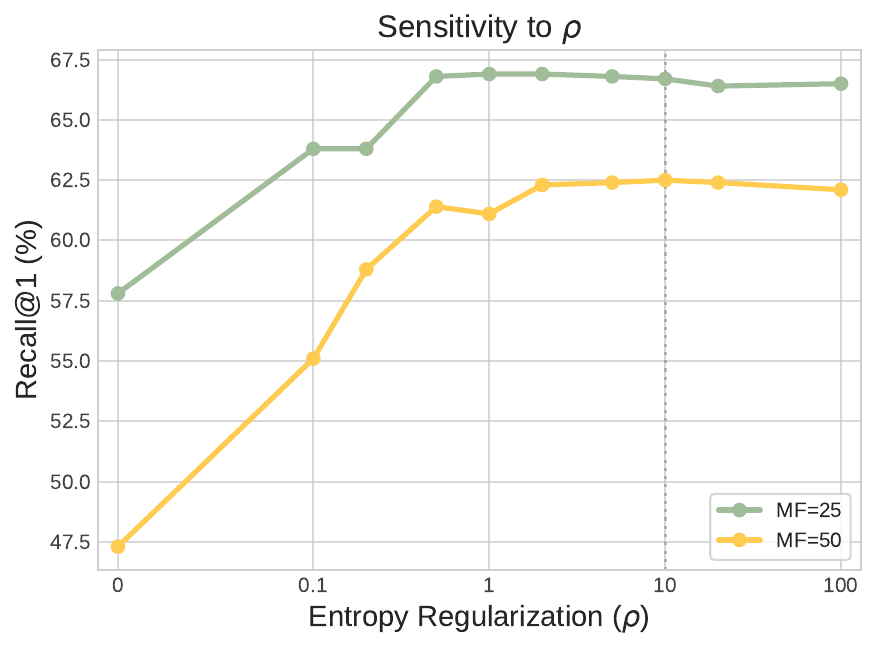}
        \caption{}
    \end{subfigure}
    \caption{\textbf{Hyperparameter Sensitivity to $\rho$.} 
     Results are obtained using ColQwen2 on the ViDoRe V1 academic tasks.}
     \label{fig:supp_hyperparam_rho}
\end{figure*}

The penalty $\rho$ controls mass conservation strictness (\cref{fig:supp_hyperparam_rho}). Although retrieval performance remains stable for $\rho \ge 1$, decreasing $\rho \to 0$ sharply degrades accuracy. This limit decouples the objective into unconstrained matching. As a result, mass concentrates on a few lowest-cost token pairs, ignoring the coverage capacity essential for visual document retrieval. Conversely, a large penalty ($\rho=100$) approximates balanced Optimal Transport (OT). This constraint prevents mass concentration and ensures broad token participation, yielding robust results. However, balanced OT strictly aligns all patches, including uninformative whitespace. Relaxing this constraint ($\rho=5$) maintains necessary coverage and allows the model to bypass uninformative regions. \cref{sec:supp_more_theory} provides a theoretical analysis of this asymptotic behavior.

\subsubsection{Efficiency of Token Compression.}
\cref{tab:indexing_time} compares the computational efficiency of token compression during offline indexing. Light-ColQwen2 employs hierarchical agglomerative clustering (HAC), which has a time complexity of $\mathcal{O}(N^2(N-K))$. Because higher compression (smaller $K$) requires more merge operations, its average indexing time increases from 196.0 ms per document (ms/doc) at MF=25 to 200.1 ms/doc at MF=50. In contrast, our method initializes the token set directly at the target size $K$ and optimizes it over $T$ Expectation-Maximization (EM) steps and $L$ Sinkhorn iterations. This yields a time complexity of $\mathcal{O}(NKDTL)$, where $D$ is the embedding dimension. Because this complexity scales linearly with $K$, our average indexing time naturally decreases from 11.4 ms/doc at MF=25 to 9.4 ms/doc at MF=50. Overall, our formulation achieves a 17$\times$--21$\times$ indexing speedup over the baseline, facilitating highly scalable large-scale document retrieval.

\begin{table}[!ht]
    \centering
    \caption{\textbf{Efficiency of Token Compression.} Average offline indexing time (ms per document) evaluated under different compression levels on the ViDoRe V1 academic tasks using ColQwen2.}
    \vspace{-0.2cm}
    \setlength{\tabcolsep}{3.4pt}
    \begin{tabular}{L{0.2\linewidth}|C{0.05\linewidth}|C{0.10\linewidth}|*{4}{C{0.096\linewidth}|}C{0.096\linewidth}}
        \toprule
        \textbf{Method} & \textbf{MF} & \textbf{ArxivQ} & \textbf{DocQ} & \textbf{InfoQ} & \textbf{TabF} & \textbf{TATQ} & \textbf{Avg.} \\
        \midrule
        \multirow{2}{*}{Light-ColQwen2} 
        & 25 & 201.6 & 203.9 & 200.3 & 167.4 & 206.6 & 196.0 \\
        & 50 & 205.4 & 208.8 & 203.8 & 171.6 & 211.2 & 200.1 \\
        \midrule
        \multirow{2}{*}{Ours}
        & 25 & 11.7 & 11.3 & 11.3 & 11.4 & 11.2 & 11.4 \\
        & 50 & 9.4 & 9.4 & 9.4 & 9.4 & 9.3 & 9.4 \\
        \bottomrule
    \end{tabular}
    \label{tab:indexing_time}
\end{table}

\section{Detailed Proofs of Theorems 1--3}
\label{sec:supp_proofs}
In this section, we restate the theorems from the main paper and provide their detailed mathematical proofs.

\subsection{Proof of MaxSim Approximation Error Bound}
\vspace*{0.15cm}
\noindent \textbf{Theorem 1 (MaxSim Approximation Error Bound).} 
\textit{Assume all token vectors in $\mathbf{Q}, \mathbf{X}, \mathbf{Y}$ are $\ell_2$-normalized. Let $\mathbf{C} \in \mathbb{R}^{N \times K}$ be the cosine-distance cost matrix where $c_{ij} = 1 - \langle \mathbf{x}_i, \mathbf{y}_j \rangle$. For a transport plan $\mathbf{P} \in \mathbb{R}^{N \times K}_+$, let $i_m^\star = \arg\max_{1 \le i \le N} \langle \mathbf{q}_m, \mathbf{x}_i \rangle$ be the best-matching original token for query token $\mathbf{q}_m$. If the row mass $r_m = \sum_{j=1}^K p_{i_m^\star j}$ is strictly positive, the MaxSim degradation is bounded by:}
\begin{equation*}
s(\mathbf{Q}, \mathbf{X}) - s(\mathbf{Q}, \mathbf{Y}) \le \sqrt{2} \sum_{m=1}^{M} \sqrt{ \frac{1}{r_m} \sum_{j=1}^{K} p_{i_m^\star j} c_{i_m^\star j} }.
\end{equation*}

\begin{proof}
For each query token $\mathbf{q}_m$, let $i_m^\star = \arg\max_i \langle \mathbf{q}_m, \mathbf{x}_i \rangle$. Defining $w_{mj} = p_{i_m^\star j}/r_m$ ensures $w_{mj} \ge 0$ and $\sum_{j=1}^K w_{mj} = 1$, as $r_m > 0$. The score gap for a single query token $\mathbf{q}_m$ is:
\begin{equation*}
\langle \mathbf{q}_m, \mathbf{x}_{i_m^\star} \rangle - \max_{1 \le j \le K} \langle \mathbf{q}_m, \mathbf{y}_j \rangle.
\end{equation*}
Bounding the maximum by the convex combination $\sum_{j=1}^K w_{mj} \langle \mathbf{q}_m, \mathbf{y}_j \rangle$ gives:
\begin{equation*}
\langle \mathbf{q}_m, \mathbf{x}_{i_m^\star} \rangle - \max_j \langle \mathbf{q}_m, \mathbf{y}_j \rangle 
\le \bigg\langle \mathbf{q}_m, \mathbf{x}_{i_m^\star} - \sum_{j=1}^K w_{mj} \mathbf{y}_j \bigg\rangle.
\end{equation*}
Applying Cauchy-Schwarz and substituting $\|\mathbf{q}_m\|_2 = 1$ yields:
\begin{equation*}
\bigg\langle \mathbf{q}_m, \mathbf{x}_{i_m^\star} - \sum_{j=1}^K w_{mj} \mathbf{y}_j \bigg\rangle 
\le \bigg\| \sum_{j=1}^K w_{mj} (\mathbf{x}_{i_m^\star} - \mathbf{y}_j) \bigg\|_2.
\end{equation*}
By the triangle inequality ($w_{mj} \ge 0$), we obtain:
\begin{equation*}
\bigg\| \sum_{j=1}^K w_{mj} (\mathbf{x}_{i_m^\star} - \mathbf{y}_j) \bigg\|_2 \le \sum_{j=1}^K w_{mj} \|\mathbf{x}_{i_m^\star} - \mathbf{y}_j\|_2.
\end{equation*}
For $\ell_2$-normalized vectors, $\|\mathbf{x}_i - \mathbf{y}_j\|_2 = \sqrt{2c_{ij}}$, leading to:
\begin{equation*}
\sum_{j=1}^K w_{mj} \|\mathbf{x}_{i_m^\star} - \mathbf{y}_j\|_2 = \sqrt{2} \sum_{j=1}^K w_{mj} \sqrt{c_{i_m^\star j}}.
\end{equation*}
Applying Jensen's inequality and summing over all $M$ queries provides:
\begin{equation*}
s(\mathbf{Q}, \mathbf{X}) - s(\mathbf{Q}, \mathbf{Y}) \le \sqrt{2} \sum_{m=1}^{M} \sqrt{ \sum_{j=1}^{K} w_{m j} c_{i_m^\star j} }.
\end{equation*}
Substituting $w_{mj} = p_{i_m^\star j}/r_m$ completes the proof.
\end{proof}

\subsection{Proof of Worst-Case Query Robustness}
To analyze the robustness of our framework against any possible unit-norm query, let $\Delta(\mathbf{X}, \mathbf{Y})$ denote the worst-case MaxSim gap for a single query token $\mathbf{q}$.

\vspace*{0.15cm}
\noindent \textbf{Theorem 2 (Worst-case Query Robustness).} 
\textit{Assume all token vectors are $\ell_2$-normalized, and let $c_{ij} = 1 - \langle \mathbf{x}_i, \mathbf{y}_j \rangle$ be the cosine cost. The worst-case MaxSim gap is bounded by the geometric covering radius:}
\begin{equation*}
\Delta(\mathbf{X}, \mathbf{Y}) := \sup_{\|\mathbf{q}\|_2=1} \left( \max_{i} \langle \mathbf{q}, \mathbf{x}_i \rangle - \max_{j} \langle \mathbf{q}, \mathbf{y}_j \rangle \right) \le \max_{i} \min_{j} \|\mathbf{x}_i - \mathbf{y}_j\|_2.
\end{equation*}
\textit{Furthermore, for any transport plan $\mathbf{P} \in \mathbb{R}^{N \times K}_+$ with row mass $r_i = \sum_{j=1}^K p_{ij} > 0$, this gap is upper-bounded by the transport cost:}
\begin{equation*}
\Delta(\mathbf{X}, \mathbf{Y}) \le \sqrt{2} \max_i \sqrt{ \frac{1}{r_i} \sum_{j=1}^K p_{ij} c_{ij} }.
\end{equation*}

\begin{proof}
For any unit vector $\mathbf{q}$, let $i^\star = \arg\max_i \langle \mathbf{q}, \mathbf{x}_i \rangle$. Since $\max_k \langle \mathbf{q}, \mathbf{y}_k \rangle \ge \langle \mathbf{q}, \mathbf{y}_j \rangle$ for any $j$, the MaxSim gap satisfies:
\begin{equation*}
\max_i \langle \mathbf{q}, \mathbf{x}_i \rangle - \max_k \langle \mathbf{q}, \mathbf{y}_k \rangle \le \langle \mathbf{q}, \mathbf{x}_{i^\star} \rangle - \langle \mathbf{q}, \mathbf{y}_j \rangle = \langle \mathbf{q}, \mathbf{x}_{i^\star} - \mathbf{y}_j \rangle.
\end{equation*}
Applying Cauchy-Schwarz and substituting $\|\mathbf{q}\|_2 = 1$ gives:
\begin{equation*}
\langle \mathbf{q}, \mathbf{x}_{i^\star} - \mathbf{y}_j \rangle 
\le \|\mathbf{x}_{i^\star} - \mathbf{y}_j\|_2.
\end{equation*}
Minimizing the right-hand side over $j$ yields:
\begin{equation*}
\max_i \langle \mathbf{q}, \mathbf{x}_i \rangle - \max_k \langle \mathbf{q}, \mathbf{y}_k \rangle 
\le \min_j \|\mathbf{x}_{i^\star} - \mathbf{y}_j\|_2 \le \max_i \min_j \|\mathbf{x}_i - \mathbf{y}_j\|_2.
\end{equation*}
Taking the supremum over all unit queries $\mathbf{q}$ yields:
\begin{equation*}
\Delta(\mathbf{X}, \mathbf{Y}) \le \max_{i} \min_{j} \|\mathbf{x}_i - \mathbf{y}_j\|_2.
\end{equation*}

To relate this to the transport plan $\mathbf{P}$, note that the minimum is upper-bounded by any convex combination. Since $p_{ij} \ge 0$, for any row $i$ with $r_i > 0$:
\begin{equation*}
\min_j c_{ij} \le \frac{1}{r_i} \sum_{j=1}^K p_{ij} c_{ij}.
\end{equation*}
Since $\|\mathbf{x}_i - \mathbf{y}_j\|_2 = \sqrt{2c_{ij}}$, we obtain:
\begin{equation*}
\min_j \|\mathbf{x}_i - \mathbf{y}_j\|_2 
= \sqrt{2 \min_j c_{ij}} 
\le \sqrt{ \frac{2}{r_i} \sum_{j=1}^K p_{ij} c_{ij} }.
\end{equation*}
Taking the maximum over $i$ yields the transport cost bound.
\end{proof}

\subsection{Proof of Ranking Stability}
To establish that bounded per-document score deviation preserves overall retrieval rankings, we consider a query $\mathbf{Q}$ and a document set $\mathcal{D}=\{1,\dots,n\}$. Let $s_d$ and $\tilde{s}_d$ denote the original and compressed scores for document $d$, respectively.

\vspace*{0.15cm}
\noindent \textbf{Theorem 3 (Ranking and Top-$k$ Stability).} 
\textit{Assume the deviation bound $|s_d - \tilde{s}_d| \le \varepsilon_d(\mathbf{Q})$ holds.}
\begin{enumerate}
\item \textit{\textbf{Pairwise Stability:} For any two documents $a$ and $b$, if their original score gap satisfies $s_a - s_b > \varepsilon_a(\mathbf{Q}) + \varepsilon_b(\mathbf{Q})$, their relative ranking is preserved after compression ($\tilde{s}_a > \tilde{s}_b$).}
\item \textit{\textbf{Top-$k$ Stability:} Let $\mathcal{T}_k$ be the set of the top-$k$ documents under the original scores. Define the top-$k$ gap as $\Delta_k(\mathbf{Q}) = s_{(k)} - s_{(k+1)}$, where $s_{(k)}$ is the $k$-th highest original score. If $\Delta_k(\mathbf{Q}) > \max_{d \in \mathcal{T}_k} \varepsilon_d(\mathbf{Q}) + \max_{d \notin \mathcal{T}_k} \varepsilon_d(\mathbf{Q})$, the top-$k$ set $\mathcal{T}_k$ is exactly preserved under the compressed scores.}
\end{enumerate}

\subsubsection{Proof of Pairwise Stability}
\begin{proof}
For any two documents $a$ and $b$, we define the original score gap as $\Delta(\mathbf{Q}) = s_a - s_b$ and the compressed score gap as $\widetilde{\Delta}(\mathbf{Q}) = \tilde{s}_a - \tilde{s}_b$. By adding and subtracting $s_a$ and $s_b$, we write:
\begin{equation*}
\widetilde{\Delta}(\mathbf{Q}) - \Delta(\mathbf{Q}) 
= (\tilde{s}_a - s_a) - (\tilde{s}_b - s_b).
\end{equation*}
Taking the absolute value and applying the triangle inequality yields:
\begin{equation*}
|\widetilde{\Delta}(\mathbf{Q}) - \Delta(\mathbf{Q})| 
\le |\tilde{s}_a - s_a| + |\tilde{s}_b - s_b| 
\le \varepsilon_a(\mathbf{Q}) + \varepsilon_b(\mathbf{Q}).
\end{equation*}
This implies the lower bound for the compressed gap:
\begin{equation*}
\widetilde{\Delta}(\mathbf{Q}) 
\ge \Delta(\mathbf{Q}) - (\varepsilon_a(\mathbf{Q}) + \varepsilon_b(\mathbf{Q})).
\end{equation*}
If the given condition $\Delta(\mathbf{Q}) > \varepsilon_a(\mathbf{Q}) + \varepsilon_b(\mathbf{Q})$ holds, then $\widetilde{\Delta}(\mathbf{Q}) > 0$. Therefore, $\tilde{s}_a > \tilde{s}_b$, proving that the pairwise ranking is strictly preserved.
\end{proof}

\subsubsection{Proof of Top-$k$ Stability}
\begin{proof}
Let $\pi$ be a permutation of $\{1, \dots, n\}$ that sorts the original scores such that $s_{\pi(1)} \ge s_{\pi(2)} \ge \dots \ge s_{\pi(n)}$, with ties broken by a fixed deterministic rule. We define the original top-$k$ index set as $\mathcal{T}_k = \{\pi(1), \dots, \pi(k)\}$. Let $\varepsilon_{\mathrm{in}}(\mathbf{Q}) = \max_{d \in \mathcal{T}_k} \varepsilon_d(\mathbf{Q})$ and $\varepsilon_{\mathrm{out}}(\mathbf{Q}) = \max_{d \notin \mathcal{T}_k} \varepsilon_d(\mathbf{Q})$. Take any document $a \in \mathcal{T}_k$ and $b \notin \mathcal{T}_k$. By the definition of the sorted scores, we have $s_a \ge s_{(k)}$ and $s_b \le s_{(k+1)}$. Thus, their original score difference is bounded by the top-$k$ gap:
\begin{equation*}
s_a - s_b \ge s_{(k)} - s_{(k+1)} = \Delta_k(\mathbf{Q}).
\end{equation*}
Using the score deviation bound $|s_d - \tilde{s}_d| \le \varepsilon_d(\mathbf{Q})$, we can lower-bound $\tilde{s}_a$ and upper-bound $\tilde{s}_b$:
\begin{equation*}
\tilde{s}_a \ge s_a - \varepsilon_a(\mathbf{Q}) \ge s_a - \varepsilon_{\mathrm{in}}(\mathbf{Q}), 
\quad \tilde{s}_b \le s_b + \varepsilon_b(\mathbf{Q}) \le s_b + \varepsilon_{\mathrm{out}}(\mathbf{Q}).
\end{equation*}
Therefore, the compressed score difference is bounded by:
\begin{equation*}
\begin{split}
\tilde{s}_a - \tilde{s}_b 
&\ge (s_a - \varepsilon_{\mathrm{in}}(\mathbf{Q})) - (s_b + \varepsilon_{\mathrm{out}}(\mathbf{Q})) \\
&= (s_a - s_b) - (\varepsilon_{\mathrm{in}}(\mathbf{Q}) + \varepsilon_{\mathrm{out}}(\mathbf{Q})) \\ 
&\ge \Delta_k(\mathbf{Q}) - (\varepsilon_{\mathrm{in}}(\mathbf{Q}) + \varepsilon_{\mathrm{out}}(\mathbf{Q})).
\end{split}
\end{equation*}
Under the condition $\Delta_k(\mathbf{Q}) > \varepsilon_{\mathrm{in}}(\mathbf{Q}) + \varepsilon_{\mathrm{out}}(\mathbf{Q})$, the right-hand side is strictly positive, meaning $\tilde{s}_a > \tilde{s}_b$ for all $a \in \mathcal{T}_k$ and $b \notin \mathcal{T}_k$. This strict separation guarantees that every top-$k$ document under the compressed scores must belong to the original set $\mathcal{T}_k$, and no document outside $\mathcal{T}_k$ can surpass them. Hence, the top-$k$ set is exactly preserved.
\end{proof}

\subsection{Linking Theorem 1 to Theorem 3}
Theorem~1 controls the score decrease $s-\tilde{s}$, whereas Theorem~3 is stated in terms of the absolute deviation $|s-\tilde{s}|$. Since the $\ell_2$-normalization in the M-step does not guarantee $\tilde{s}\le s$, we next derive auxiliary results that bound the reverse direction as well.

\vspace*{0.15cm}

\begin{proposition}[Closed-form M-step on the unit sphere]
Fix $\mathbf{P}\in\mathbb{R}_+^{N\times K}$ and assume $\|\mathbf{x}_i\|_2=1$ for all $i$. For each anchor $\mathbf{y}_j$ under the constraint $\|\mathbf{y}_j\|_2=1$, consider
\begin{equation*}
\min_{\|\mathbf{y}_j\|_2=1}\sum_{i=1}^N p_{ij}\bigl(1-\langle \mathbf{x}_i,\mathbf{y}_j\rangle\bigr).
\end{equation*}
If $\mathbf{v}_j := \sum_{i=1}^N p_{ij}\mathbf{x}_i \neq \mathbf{0}$, the unique minimizer is
\begin{equation*}
\mathbf{y}_j^\star
= \frac{\sum_{i=1}^N p_{ij}\mathbf{x}_i}{\left\|\sum_{i=1}^N p_{ij}\mathbf{x}_i\right\|_2}
= \frac{\mathbf{v}_j}{\|\mathbf{v}_j\|_2}.
\end{equation*}
\end{proposition}

\vspace*{0.1cm}
\begin{proof}
The objective can be written as $\sum_{i=1}^N p_{ij} - \left\langle \sum_{i=1}^N p_{ij}\mathbf{x}_i,\mathbf{y}_j \right\rangle = \sum_{i=1}^N p_{ij} - \langle \mathbf{v}_j,\mathbf{y}_j\rangle$.
Since the first term is constant in $\mathbf{y}_j$, minimizing the objective is equivalent to maximizing
$\langle \mathbf{v}_j,\mathbf{y}_j\rangle$ subject to $\|\mathbf{y}_j\|_2=1$. By Cauchy--Schwarz,
\begin{equation*}
\langle \mathbf{v}_j,\mathbf{y}_j\rangle
\le \|\mathbf{v}_j\|_2\|\mathbf{y}_j\|_2
= \|\mathbf{v}_j\|_2,
\end{equation*}
with equality if and only if $\mathbf{y}_j=\mathbf{v}_j/\|\mathbf{v}_j\|_2$.
\end{proof}

\vspace*{0.15cm}
\begin{lemma}[Column-wise cost identity under the optimal M-step]
\label{lem:columwise_cost}
Under the assumptions of Proposition~1, define $\kappa_j := \sum_{i=1}^N p_{ij}$ and $c_{ij} := 1-\langle \mathbf{x}_i,\mathbf{y}_j^\star\rangle$. If $\mathbf{v}_j \neq \mathbf{0}$, then
\begin{equation*}
\frac{1}{\kappa_j}\sum_{i=1}^N p_{ij}c_{ij} = 1-\frac{\|\mathbf{v}_j\|_2}{\kappa_j}.
\end{equation*}
Equivalently,
\begin{equation*}
\|\mathbf{v}_j\|_2 = \kappa_j-\sum_{i=1}^N p_{ij}c_{ij}.
\end{equation*}
\end{lemma}

\vspace*{0.1cm}
\begin{proof}
By Proposition~1, $\mathbf{y}_j^\star = \mathbf{v}_j/\|\mathbf{v}_j\|_2$. Therefore,
\begin{equation*}
\|\mathbf{v}_j\|_2 
= \langle \mathbf{y}_j^\star,\mathbf{v}_j \rangle
= \sum_{i=1}^N p_{ij}\langle \mathbf{y}_j^\star,\mathbf{x}_i\rangle
= \sum_{i=1}^N p_{ij}(1-c_{ij})
= \kappa_j-\sum_{i=1}^N p_{ij}c_{ij}.
\end{equation*}
Rearranging completes the proof.
\end{proof}

\begin{lemma}[Reverse MaxSim bound under the optimal M-step]
\label{lem:reveerse_bound}
Assume all token vectors in $\mathbf{Q},\mathbf{X},\mathbf{Y}$ are $\ell_2$-normalized, and let $\mathbf{Y}$ be obtained by the optimal M-step in Proposition~1. For each query token $\mathbf{q}_m$, let $j_m^\star \in \arg\max_{1\le j\le K}\langle \mathbf{q}_m,\mathbf{y}_j\rangle$ and define the selected column mass $\kappa_m := \sum_{i=1}^N p_{i j_m^\star}$. If $\sum_{i=1}^N p_{i j_m^\star}\mathbf{x}_i \neq \mathbf{0}$, then
\begin{equation*} 
\max_{1\le j\le K}\langle \mathbf{q}_m,\mathbf{y}_j\rangle - \max_{1\le i\le N}\langle \mathbf{q}_m,\mathbf{x}_i\rangle 
\le \frac{1}{\kappa_m}\sum_{i=1}^N p_{i j_m^\star}c_{i j_m^\star}. 
\end{equation*} 
Consequently,
\begin{equation*} 
s(\mathbf{Q},\mathbf{Y})-s(\mathbf{Q},\mathbf{X}) 
\le \sum_{m=1}^M \frac{1}{\kappa_m}\sum_{i=1}^N p_{i j_m^\star}c_{i j_m^\star}. 
\end{equation*}
\end{lemma}

\vspace*{0.1cm}
\begin{proof}
Fix a query token $\mathbf{q}_m$ and write $j^\star = j_m^\star$. Noting that $\kappa_m > 0$ (as $\sum_{i=1}^N p_{i j^\star}\mathbf{x}_i \neq \mathbf{0}$), we define
\begin{equation*}
\bar w_{mi} := \frac{p_{i j^\star}}{\kappa_m},
\qquad \bar w_{mi}\ge 0,
\qquad \sum_{i=1}^N \bar w_{mi}=1.
\end{equation*}
By the definition of $j^\star$, the score gap for $\mathbf{q}_m$ simplifies to $\langle \mathbf{q}_m, \mathbf{y}_{j^\star} \rangle - \max_i \langle \mathbf{q}_m, \mathbf{x}_i \rangle$. Bounding the maximum over $i$ by the convex combination gives:
\begin{equation*}
\langle \mathbf{q}_m,\mathbf{y}_{j^\star}\rangle - \max_i\langle \mathbf{q}_m,\mathbf{x}_i\rangle 
\le \bigg\langle \mathbf{q}_m,\, \mathbf{y}_{j^\star}-\sum_{i=1}^N \bar w_{mi}\mathbf{x}_i \bigg\rangle.
\end{equation*}
Applying Cauchy--Schwarz and using $\|\mathbf{q}_m\|_2=1$, we obtain:
\begin{equation*}
\langle \mathbf{q}_m,\mathbf{y}_{j^\star}\rangle - \max_i\langle \mathbf{q}_m,\mathbf{x}_i\rangle 
\le \bigg\| \mathbf{y}_{j^\star}-\sum_{i=1}^N \bar w_{mi}\mathbf{x}_i \bigg\|_2.
\end{equation*}

Next, let $\mathbf{v}_{j^\star}:=\sum_{i=1}^N p_{i j^\star}\mathbf{x}_i$. By Proposition~1, $\mathbf{y}_{j^\star}=\mathbf{v}_{j^\star}/\|\mathbf{v}_{j^\star}\|_2$ and $\sum_{i=1}^N \bar w_{mi}\mathbf{x}_i=\mathbf{v}_{j^\star}/\kappa_m$. Since these two vectors are collinear and point in the same direction, the norm of their difference reduces to:
\begin{equation*}
\bigg\| \mathbf{y}_{j^\star}-\sum_{i=1}^N \bar w_{mi}\mathbf{x}_i \bigg\|_2 
= \left|1-\frac{\|\mathbf{v}_{j^\star}\|_2}{\kappa_m}\right|.
\end{equation*}
By the triangle inequality and $\|\mathbf{x}_i\|_2=1$, we have:
\begin{equation*}
\|\mathbf{v}_{j^\star}\|_2 
= \bigg\|\sum_{i=1}^N p_{i j^\star}\mathbf{x}_i\bigg\|_2 
\le \sum_{i=1}^N p_{ij^\star}\|\mathbf{x}_i\|_2 = \kappa_m.
\end{equation*}
This guarantees that $1-\frac{\|\mathbf{v}_{j^\star}\|_2}{\kappa_m} \ge 0$, allowing us to remove the absolute value. Substituting the identity from \cref{lem:columwise_cost} ($1-\frac{\|\mathbf{v}_{j^\star}\|_2}{\kappa_m} = \frac{1}{\kappa_m}\sum_{i=1}^N p_{i j^\star}c_{i j^\star}$) into the inequality chain yields:
\begin{equation*}
\max_j\langle \mathbf{q}_m,\mathbf{y}_j\rangle - \max_i\langle \mathbf{q}_m,\mathbf{x}_i\rangle 
\le \frac{1}{\kappa_m}\sum_{i=1}^N p_{i j_m^\star}c_{i j_m^\star}.
\end{equation*}
Summing over $m=1,\dots,M$ completes the proof.
\end{proof}

\vspace*{0.15cm}
\begin{corollary}[Two-sided score deviation bound]
\label{cor:twoside_bound}
Under the assumptions of Theorem~1 and Lemma~2, define
\begin{equation*}
\varepsilon^{\downarrow}(\mathbf{Q}) 
:= \sqrt{2}\sum_{m=1}^{M}\sqrt{\frac{1}{r_m}\sum_{j=1}^{K} p_{i_m^\star j} c_{i_m^\star j}},
\end{equation*}
and
\begin{equation*}
\varepsilon^{\uparrow}(\mathbf{Q})
:= \sum_{m=1}^{M}\frac{1}{\kappa_m}\sum_{i=1}^{N} p_{i j_m^\star} c_{i j_m^\star}.
\end{equation*}
Then
\begin{equation*}
|s(\mathbf{Q},\mathbf{X})-s(\mathbf{Q},\mathbf{Y})| 
\le \max\{\varepsilon^{\downarrow}(\mathbf{Q}),\varepsilon^{\uparrow}(\mathbf{Q})\}.
\end{equation*}
\end{corollary}

\begin{proof}
Theorem~1 gives
\begin{equation*}
s(\mathbf{Q},\mathbf{X})-s(\mathbf{Q},\mathbf{Y}) 
\le \varepsilon^{\downarrow}(\mathbf{Q}),
\end{equation*}
while \cref{lem:reveerse_bound} gives
\begin{equation*}
s(\mathbf{Q},\mathbf{Y})-s(\mathbf{Q},\mathbf{X}) 
\le \varepsilon^{\uparrow}(\mathbf{Q}).
\end{equation*}
Hence both directions are bounded, which implies
\begin{equation*}
|s(\mathbf{Q},\mathbf{X})-s(\mathbf{Q},\mathbf{Y})| 
\le \max\{\varepsilon^{\downarrow}(\mathbf{Q}),\varepsilon^{\uparrow}(\mathbf{Q})\}.
\end{equation*}
\end{proof}

\noindent \textbf{Remark on Ranking Stability.}
The results above bridge the gap between Theorem~1 and Theorem~3. Applying \cref{cor:twoside_bound} to each document $d$, yields a document-specific deviation bound $\varepsilon_d(\mathbf{Q}) := \max\{\varepsilon_d^{\downarrow}(\mathbf{Q}), \varepsilon_d^{\uparrow}(\mathbf{Q})\}$. Substituting this bound into Theorem~3 yields a sufficient condition under which our OT-based compression preserves pairwise rankings and the top-$k$ set.

\section{Additional Theoretical Analysis}
\label{sec:supp_more_theory}

\subsection{Asymptotic Behavior of the Marginal Relaxation}
To analyze the asymptotic behavior of marginal relaxation, we define the UOT objective function $F_\rho(\mathbf{P})$ as:
\begin{equation}
\label{eq:f_rho}
F_\rho(\mathbf{P}) := \langle \mathbf{P}, \mathbf{C}\rangle_{F} - \gamma H(\mathbf{P}) + \rho \, G(\mathbf{P}),
\end{equation}
where $G(\mathbf{P}) := \mathrm{KL}(\mathbf{P}\mathbf{1}\,\|\,\boldsymbol{\mu}) + \mathrm{KL}(\mathbf{P}^\top\mathbf{1}\,\|\,\boldsymbol{\nu})$ represents the marginal penalty terms, and $\mathbf{1}$ denotes an all-ones vector of appropriate dimension. The term $F_0(\mathbf{P})$ denotes the objective without the marginal constraints:
\begin{equation}
\label{eq:f_zero}
F_0(\mathbf{P}) := \langle \mathbf{P}, \mathbf{C}\rangle_{F} - \gamma H(\mathbf{P}).
\end{equation}

\begin{proposition}[Asymptotic limits of $\rho$ via accumulation points]
\label{prop:rho_app}
Fix $\gamma>0$ and let $\mathbf{P}_\rho\in\arg\min_{\mathbf{P}\ge 0} F_\rho(\mathbf{P})$.
\begin{enumerate}
\item \textbf{Limit $\rho\to 0$:} For any sequence $\rho_n\downarrow 0$, every convergent subsequence $\mathbf{P}_{\rho_{n_k}}\to\bar{\mathbf{P}}$ satisfies
\begin{equation*}
\bar{\mathbf{P}}\in\arg\min_{\mathbf{P}\ge 0} F_0(\mathbf{P}).
\end{equation*}

\item \textbf{Limit $\rho\to\infty$:} For any sequence $\rho_n\to\infty$, every convergent subsequence $\mathbf{P}_{\rho_{n_k}}\to\bar{\mathbf{P}}$ satisfies $\bar{\mathbf{P}}\in\mathcal{P}(\boldsymbol{\mu},\boldsymbol{\nu})$ and
\begin{equation*}
\bar{\mathbf{P}}\in\arg\min_{\mathbf{P}\in\mathcal{P}(\boldsymbol{\mu},\boldsymbol{\nu})} F_0(\mathbf{P}).
\end{equation*}
\end{enumerate}
\end{proposition}

\begin{proof}
\textbf{(1) Case $\rho\to 0$:}
Fix any sequence $\rho_n\downarrow 0$ and let $\mathbf{P}_n:=\mathbf{P}_{\rho_n}$. Assume $\mathbf{P}_{n_k}\to \bar{\mathbf{P}}$ along a convergent subsequence. For any fixed $\mathbf{P}\ge 0$, the optimality of $\mathbf{P}_n$ ensures:
\begin{equation*}
F_{\rho_n}(\mathbf{P}_n)\le F_{\rho_n}(\mathbf{P})=F_0(\mathbf{P})+\rho_n G(\mathbf{P}),
\end{equation*}
where $G(\mathbf{P}):=\mathrm{KL}(\mathbf{P}\mathbf{1}\|\boldsymbol{\mu})+\mathrm{KL}(\mathbf{P}^\top\mathbf{1}\|\boldsymbol{\nu})\ge 0$. Since $F_{\rho_n}(\mathbf{P}_n)=F_0(\mathbf{P}_n)+\rho_n G(\mathbf{P}_n)\ge F_0(\mathbf{P}_n)$, we obtain:
\begin{equation*}
F_0(\mathbf{P}_n)\le F_0(\mathbf{P})+\rho_n G(\mathbf{P}).
\end{equation*}
Taking $\limsup_{n\to\infty}$ yields 
\begin{equation*}
\limsup_{n\to\infty} F_0(\mathbf{P}_n)\le F_0(\mathbf{P}).
\end{equation*} 
By the continuity of $F_0$ in finite dimensions and $\mathbf{P}_{n_k}\to\bar{\mathbf{P}}$, we have $F_0(\bar{\mathbf{P}})=\lim_{k\to\infty}F_0(\mathbf{P}_{n_k})\le \limsup_{n\to\infty}F_0(\mathbf{P}_n)$. Hence, $F_0(\bar{\mathbf{P}})\le F_0(\mathbf{P})$ for every $\mathbf{P}\ge 0$, establishing that $\bar{\mathbf{P}}\in\arg\min_{\mathbf{P}\ge 0}F_0(\mathbf{P})$.

\medskip
\textbf{(2) Case $\rho\to\infty$:}
Fix any sequence $\rho_n\to\infty$ and let $\mathbf{P}_n:=\mathbf{P}_{\rho_n}$. Choose a feasible plan $\mathbf{P}^0\in\mathcal{P}(\boldsymbol{\mu},\boldsymbol{\nu})$ (e.g., $\mathbf{P}^0_{ij}=\mu_i\nu_j$), so that $G(\mathbf{P}^0)=0$ and $F_{\rho_n}(\mathbf{P}^0)=F_0(\mathbf{P}^0)$ for all $n$. By optimality:
\begin{equation*}
F_0(\mathbf{P}_n)+\rho_n G(\mathbf{P}_n) = F_{\rho_n}(\mathbf{P}_n) \le F_{\rho_n}(\mathbf{P}^0) = F_0(\mathbf{P}^0).
\end{equation*}
Rearranging this gives:
\begin{equation}
\label{eq:penalty_goes_zero}
\rho_n G(\mathbf{P}_n)\le F_0(\mathbf{P}^0)-F_0(\mathbf{P}_n)\le F_0(\mathbf{P}^0)-\inf_{\mathbf{P}\ge 0}F_0(\mathbf{P})<\infty.
\end{equation}

Indeed, let $\mathbf{r}_n:=\mathbf{P}_n\mathbf{1}\in\mathbb{R}^N_+$. From \eqref{eq:penalty_goes_zero} and $G(\mathbf{P}_n)\ge \mathrm{KL}(\mathbf{r}_n\|\boldsymbol{\mu})\ge 0$, it follows that $\mathrm{KL}(\mathbf{r}_n\|\boldsymbol{\mu})\le F_0(\mathbf{P}^0)/\rho_n\to 0$. By Lemma~\ref{lem:KL_mass}, this implies the total mass $m_n:=\mathbf{1}^\top \mathbf{r}_n\to 1$. Similarly, with $\mathbf{c}_n:=\mathbf{P}_n^\top\mathbf{1}$, we have $\mathrm{KL}(\mathbf{c}_n\|\boldsymbol{\nu})\to 0$ and $\mathbf{1}^\top \mathbf{c}_n\to 1$.

For any convergent subsequence $\mathbf{P}_{n_k}\to\bar{\mathbf{P}}$, the continuity of matrix multiplication and the KL divergence lower bound guarantee that $\mathrm{KL}(\mathbf{P}_{n_k}\mathbf{1}\|\boldsymbol{\mu})\to 0$ implies $\mathbf{P}_{n_k}\mathbf{1}\to \boldsymbol{\mu}$, and similarly $\mathbf{P}_{n_k}^\top\mathbf{1}\to \boldsymbol{\nu}$. Therefore, $\bar{\mathbf{P}}\in\mathcal{P}(\boldsymbol{\mu},\boldsymbol{\nu})$.

Finally, for any $\mathbf{P}\in\mathcal{P}(\boldsymbol{\mu},\boldsymbol{\nu})$, we have $G(\mathbf{P})=0$, leading to $F_{\rho_n}(\mathbf{P})=F_0(\mathbf{P})$. Optimality implies $F_0(\mathbf{P}_n)+\rho_n G(\mathbf{P}_n)\le F_0(\mathbf{P})$. Letting $n=n_k\to\infty$ along the convergent subsequence and utilizing $G(\mathbf{P}_{n_k})\to 0$ yields $F_0(\bar{\mathbf{P}})\le F_0(\mathbf{P})$ for all $\mathbf{P}\in\mathcal{P}(\boldsymbol{\mu},\boldsymbol{\nu})$. Thus, $\bar{\mathbf{P}}\in\arg\min_{\mathbf{P}\in\mathcal{P}(\boldsymbol{\mu},\boldsymbol{\nu})}F_0(\mathbf{P})$.
\end{proof}

In the proof of Proposition~\ref{prop:rho_app}, we use the following standard consequence of the log-sum inequality to bound the total mass.

\begin{lemma}[KL Divergence Mass Bound]
\label{lem:KL_mass}
Let $\mathbf{a},\mathbf{b}\in\mathbb{R}^n_{+}$ with $\mathbf{1}^\top \mathbf{b}=1$, and define the total mass $m:=\mathbf{1}^\top \mathbf{a}$. Then,
\begin{equation*}
\mathrm{KL}(\mathbf{a}\|\mathbf{b}) \ge m\log m - m + 1.
\end{equation*}
In particular, if $\mathrm{KL}(\mathbf{a}\|\mathbf{b})\to 0$, then $m\to 1$.
\end{lemma}

\begin{proof}
By the log-sum inequality,
\begin{equation*}
\sum_{i=1}^n a_i\log\frac{a_i}{b_i} \ge \left(\sum_{i=1}^n a_i\right)\log\frac{\sum_{i=1}^n a_i}{\sum_{i=1}^n b_i} = m\log m,
\end{equation*}
since $\sum_i b_i=1$. Using the definition $\mathrm{KL}(\mathbf{a}\|\mathbf{b})=\sum_i a_i\log(a_i/b_i)-\sum_i a_i+\sum_i b_i$, we obtain:
\begin{equation*}
\mathrm{KL}(\mathbf{a}\|\mathbf{b}) \ge m\log m - m + 1.
\end{equation*}
The function $m\mapsto m\log m - m + 1$ is non-negative and vanishes strictly at $m=1$. Hence, $\mathrm{KL}(\mathbf{a}\|\mathbf{b})\to 0$ implies $m\to 1$.
\end{proof}

\bibliographystyle{splncs04}